\documentclass[10pt,journal,twoside,twocolumn]{IEEEtran} %default for journal is also 10pt (also requirement for TIP)
\usepackage{cite}
\ifCLASSINFOpdf
\else
\fi
\usepackage{graphicx}
\usepackage[table,xcdraw]{xcolor}
\usepackage{algorithm,algpseudocode}% http://ctan.org/pkg/algorithmicx
\usepackage{amsmath,amsfonts,amssymb}
\usepackage{float}
\usepackage{hyperref}
\usepackage[inline]{enumitem}

\usepackage{multirow}
\usepackage{hhline}

\usepackage{soul}

\algnewcommand{\LineComment}[1]{\State \(\triangleright\) #1}
\usepackage{comment}

\let\oldnl\nl% Store \nl in \oldnl
\newcommand{\nonl}{\renewcommand{\nl}{\let\nl\oldnl}}% Remove line number for one line
\usepackage{relsize}
\usepackage{pifont}% http://ctan.org/pkg/pifont
\newcommand{\cmark}{\ding{51}}%
\newcommand{\xmark}{\ding{55}}%
\usepackage{xcolor}

\renewcommand{\vec}[1]{\mathbf{#1}}
\usepackage{amsmath}
\usepackage{mathtools}
\newif\ifmark

\newlength\myindent
\makeatletter
\def\ps@IEEEtitlepagestyle{
  \def\@oddfoot{\mycopyrightnotice}
  \def\@evenfoot{}
}
\def\mycopyrightnotice{
  {\footnotesize
  \begin{minipage}{\textwidth}
  ~\copyright~2021 IEEE. Personal use of this material is permitted. Permission from IEEE must be obtained for all other uses, in any current or future media, including reprinting/republishing this material for advertising or promotional purposes, creating new
collective works, for resale or redistribution to servers or lists, or reuse of any copyrighted component of this work in other works. DOI: 10.1109/TIP.2021.3127847
  \end{minipage}
  }
}

\begin{document}
\marktrue %%use this is you want to mark the text
% \markfalse %%use this is you want to unmark the
%old title ---->
%\title{D\'{e}j\`{a} vu from the SVM Era: Example-based Explanation  with Outlier Detection for Image Classification}
%\title{Example-based Explanation and Detection of Outliers via  Prototypical Examples for Image Classification}
\title{Toward Scalable and Unified Example-based Explanation and Outlier Detection}

\author{Penny~Chong, 
        Ngai-Man Cheung,
        Yuval~Elovici,
        and~Alexander~Binder% <-this % stops a space
\thanks{P. Chong and N.-M. Cheung are with the Information Systems Technology and Design Pillar, Singapore University of Technology and Design, 487372, Singapore. (e-mail: penny\_chong@mymail.sutd.edu.sg; ngaiman\_cheung@sutd.edu.sg)}
\thanks{Y. Elovici is with the Department of Software and Information Systems Engineering, Ben-Gurion University of the Negev, Beer-Sheva, 84105, Israel. (email: elovici@bgu.ac.il)}
\thanks{A. Binder is with the Department of Informatics, University of Oslo, 0373 Oslo, Norway. (email: alexabin@ifi.uio.no)}
%\thanks{M. Shell was with the Department of Electrical and Computer Engineering, Georgia Institute of Technology, Atlanta,GA, 30332 USA e-mail: (see http://www.michaelshell.org/contact.html).}% <-this % stops a space
}

\maketitle

\begin{abstract}
%The lack of transparency in the inference process of deep neural networks has become a concern in the recent years. 
%Due to the risks involved in high-stakes tasks, it is important for neural networks to provide predictions together with explanations to understand the features that have contributed to the decision. 
When neural networks are employed for high-stakes decision-making, it is desirable that they provide explanations for their prediction in order for us to understand the features that have contributed to the decision. At the same time, it is important to flag potential outliers for in-depth verification by domain experts.
%Therefore, the explainable AI area has gained significant interest from researchers attempting to address the lack of transparency issue in deep neural networks. 
In this work we propose to unify two differing aspects of explainability with outlier detection.
%In this work, we address the transparency issue in prediction by proposing a two-fold explanation. 
We argue for a broader adoption of prototype-based student networks capable of providing an example-based explanation for their prediction and at the same time identify regions of similarity between the predicted sample and the examples. The examples are real prototypical cases sampled from the training set via our novel iterative prototype replacement algorithm. % to give a meaningful explanation. 
%Besides explainability, the other aspect we address in this work is the use of prototype similarity scores in identifying outliers for neural networks deployed in high-stakes settings. 
 Furthermore, we propose to use the prototype similarity scores for identifying outliers.
%In addition, we evaluate the reliability of our proposed network in identifying outliers by using the prototype similarity scores. 
We compare performances in terms of the classification, explanation quality, and outlier detection of our proposed network with other baselines. We show that our prototype-based networks beyond similarity kernels deliver meaningful explanations and promising outlier detection results without compromising classification accuracy.

\end{abstract}


% \fi
\IEEEpeerreviewmaketitle

\section{Introduction}
\IEEEPARstart{D}{eep} neural networks (DNNs) are widely used in various fields because they show superior performance. Explainable AI, aiming at transparency for predictions \cite{simonyan2013deep,zeiler2014visualizing,montavon2019layer} has gained research attention in recent years.  
%In this work, we revisit explainability by examples. 
Since humans are known to generalize fast from a few examples, it is suitable to explain a prediction of a test sample via a set of similar examples.
Such an example-based explanation can simply be achieved using a similarity search over an available dataset based on a metric defined by the feature maps obtained from the trained model. 
To achieve this, one natural option is to search in an available dataset for the nearest candidates to the test sample in the feature map space. 
%For this, one computes a feature map for the test sample to be explained, and then searches an available dataset in the same feature map space for the nearest candidates to the test sample. 
This yields an informative visualization of the feature embedding learned by the model, yet the similar examples found do not participate in the prediction of the test sample. Even if these similar examples share the same prediction with the test sample and are close in feature space, it is not obvious to what \textit{quantifiable extent} the model shares the same reasoning between examples and the test sample, and which parts of the test sample and nearest neighbors share the same features. % to predict the same label. %\footnote{To see this one can consider the sparsity of dual kernel based predictors as an argument.} 
An alternative to achieve such explanation by examples is to employ kernel-based predictors \cite{cortes1995support}. %SVM, Kernel perceptron, kernel ridge regression
These methods compute a weighted sum of similarities between the test and the training samples, and thus the impact of each training sample on the prediction of the test sample is naturally quantifiable.
Although neural networks perform very well these days without the need for kernel-based setups, one may consider training a prototype-based student network from an arbitrarily structured teacher network to provide an example-based explanation\footnote{The teacher network is a typical convolutional neural network (CNN) for image classification tasks. The soft labels obtained from the teacher will be used for student-teacher learning to train a prototype-based student network.}. The prototypes in the student network participate in the network prediction, which resembles the participation of training samples in kernel-based predictions.

In this work we propose to unify two aspects of explanation with outlier detection, independent of the structure of the original model, by employing a prototype-based student network to address  the three goals.
%In this paper, we propose a prototype-based student network which provides prediction explanation based on real examples selected from the training set via our novel iterative prototype replacement algorithm. %auxiliary output branch. 
The explanation for predictions is given by the network in these two aspects:
i) the top-$k$ most similar prototypical examples and
ii) regions of similarity between the prediction sample and the top-$k$ prototypes.
Contrasting previous prototype learning approaches \cite{yang2018robust,li2017deep,ming2019interpretable,chen2019looks,hase2019interpretable,rymarczyk2020protopshare,nauta2020neural}, we refrain from directly training prototype vectors in the latent space to avoid reconstruction errors when visualizing the learned prototype vectors in the input space. 
In lieu of this, we introduce an auxiliary output branch in the network for iterative prototype replacement along the lines of prototype selection methods \cite{gurumoorthy2019efficient, kim2016examples, arik2020protoattend} to select representative training examples for prediction inspired by dual kernel support vector machines (SVMs) \cite{cortes1995support}. Our method omits
the need to map or decode prototype vectors and guarantees an
explanation relative to real examples present in the training distribution. The prototype importance weight learned via the prototype replacement auxiliary loss also enables the pruning of uninformative prototypes.% to reduce the size of the neural network.

With the proposed approach, it is imperative to consider the question of performance tradeoff when converting a standard CNN teacher network into a prototype-based student network. We quantify the performance of the predictors in the following three aspects: the prediction accuracy, the quality of explanation of the student network compared to the teacher network, and finally the competence of each predictor in quantifying the outlierness of a given sample. The first two aspects are the natural tradeoffs to be considered when training any surrogate model, while the last aspect measures the outlier sensitivity of predictors in high-stakes environments. %beyond the usual scientific benchmarks. 
It is necessary for predictors deployed in high-stakes settings to be equipped with the capability to flag inputs that are either anomalous or poorly represented by the training set for additional validation by experts. Prototype-based approaches deliver this property naturally by the sequence of sorted similarities between the test sample and the prototypes even though they are not trained primarily for outlier detection. The following summarizes our contribution in this work:
\begin{enumerate}
\item We argue for unified interpretable models capable of explaining and detecting outliers with prototypical examples. 
We demonstrate that a student network based on prototypes is capable of performing two types of explanation tasks and one outlier task simultaneously. We select prototypes from the training set that guarantee that the explanation based on examples comes from the true data distribution. 

The prototype network provides two layers of explanation, first by identifying the top-$k$ nearest prototypical examples, and second, by showing the pixel evidence of similarity between each of these examples and the test sample. The latter is achieved by backpropagating scores from similarity layers using the Layer-wise Relevance Propagation (LRP) approach for explainability.
\item We introduce a novel iterative prototype replacement algorithm relying on training with masks and a widely reusable auxiliary loss term. We demonstrate that this prototype replacement approach scales to larger datasets such as LSUN and PCam.

\item We propose various generic head architectures that compute prototype-based predictions from a generic CNN feature extractor and evaluate their prediction accuracy, explanation quality, and outlier detection performance. 
% \item We propose and analyze various generic head architectures which compute prototype-based predictions from a generic CNN feature extractor, with respect to predictive accuracy, explanation quality and outlier detection performance. 
%\item We compare for the first time the pixelwise explanation quality between a typical teacher architecture and the student architectures defined by their heads. 

\item We revisit a method to quantify the degree of outlierness for a sample 
based on the sorted prototype similarity scores.
By doing so, we avoid the complex issue of hyperparameter selection in unsupervised outlier detection methods without resorting to problematic tuning of test sets. A smaller contribution lies in the derivation of the prototype similarity scores depending on the network architectures.
\end{enumerate}

\section{Related Work}
\textbf{Prototype Learning}
%This work draws from the literature on prototype learning. 
One of the earliest works in prototype learning is the k-nearest neighbor (k-NN) classifier \cite{altman1992introduction}.
 In recent years, researchers have proposed combining DNNs with prototype-based classifiers. The work \cite{yang2018robust} proposed a framework known as  convolutional prototype learning (CPL), where the prototype vectors are optimized to encourage representations that are intraclass compact and interclass separable. Subsequently, interpretable prototype networks \cite{li2017deep,ming2019interpretable} were proposed to provide explanations based on similar prototypes for image and sequence classification tasks. The former used a decoder, while the latter projected each prototype to its closest embedding from the training set. Another work \cite{chen2019looks} introduced a prototypical part network (ProtoPNet) that is able to present prototypical cases that are similar to the parts the network looks at. These explanations are used during classification and not created post hoc. HPnet \cite{hase2019interpretable} classifies objects using hierarchically organized prototypes that are predefined in a taxonomy and is a generalization of ProtoPNet.  Explanation-by-examples such as the use of prototypical examples is often preferred over  superimposition explanations \cite{chattopadhay2018grad,simonyan2013deep,lundberg2017unified} for nontechnical end-users \cite{jeyakumar2020can}. Further works in this domain include 
 \cite{rymarczyk2020protopshare,nauta2020neural}.

In contrast to the aforementioned methods, we select prototypes directly from the training set and do not use trainable prototype vectors. Related works on prototype selection are \cite{gurumoorthy2019efficient, kim2016examples}, which involve mining representative samples such as prototypes and criticism samples to provide an optimally compact representation of the underlying data distribution.
 Our work is related to ProtoAttend \cite{arik2020protoattend}, which selects  input-dependent prototypes  via an attention mechanism applied between the input and the database of prototype candidates. A notable difference between our work and ProtoAttend is that our prototypes are global prototypes (fixed after network convergence), whereas they learn input-dependent prototypes that require a sufficiently large prototype candidate database and high memory consumption during training to obtain good performance \cite{arik2020protoattend}. %Since their set of prototypes is not fixed, the keys and values from the prototype candidate database also need to be precomputed for inference to reduce the high overhead. Our approach which selects global prototypes does not have this overhead during inference.
In terms of explanation, our method is related to \cite{chen2019looks,hase2019interpretable,rymarczyk2020protopshare,nauta2020neural}. These methods explicitly focused on a patchwise representation of prototypes to deliver prototypical part explanation, whereas in our approach, a prototype represents the entire sample and we present a more fine-grained explanation in terms of pixel similarity between the prototype and test sample to explain their respective contribution to the membership of the predicted class using a post hoc explainability method \cite{montavon2019layer}.
%Several other works use prototype-based predictors for few-shot learning \cite{snell2017prototypical,luattentive}.

\textbf{Outlier Detection.}
We revisit the idea of quantifying outliers %, also known as out-of-distribution samples,
using statistics from the set of top-$k$ similarity scores. For this, k-NN-based distance methods have been frequently considered in the form of summed distances \cite{10.1007/3-540-45681-3_2}, local outlier factors \cite{10.1145/342009.335388,DBLP:conf/mldm/LateckiLP07}, combinations of distance and density estimates \cite{kim2006prototype}, or kernel-based statistics \cite{10.1007/978-3-642-13062-5_16}, and such methods are still of recent interest \cite{NIPS2019_9274} due to favorable theoretical properties.  Unlike these works, we are interested in the performance of similarities that do not define a kernel, but rather emerge due to attention-based operations on feature maps. % and we omit to compute density estimates of nearest neighbors of a test sample.
A simple baseline for out-of-distribution detection is the maximum softmax probability method \cite{hendrycks2016baseline} which assumes that erroneously classified samples or outliers exhibit lower probability scores for the predicted class than the correctly classified example. Several other works related to the detection of out-of-distribution samples are \cite{liang2017enhancing,hsu2020generalized,hendrycks2018deep}. Generalized ODIN \cite{hsu2020generalized} improved upon the ODIN \cite{liang2017enhancing} method, which is based on temperature scaling and adding small perturbations, by proposing to decompose confidence scores and a modified input preprocessing method that eliminates the need to tune hyperparameters with out-of-distribution samples.
%For comparison, we employ the baseline method \cite{hendrycks2016baseline} for the other predictors and isolation forest (IF) \cite{liu2008isolation} to evaluate the effectiveness of the prototype similarity scores derived from our student architectures. 
Our formulation of the outlier score using the sum of prototype similarity scores is based on the idea in \cite{10.1007/978-3-642-13062-5_16} and is also the generalization of the confidence score in ProtoAttend \cite{arik2020protoattend}.

\section{Methodology}
In a practical development task, many researchers prefer to start with a lower complexity baseline. As of 2021, this baseline for image classification  would still be the conventional CNN  for many practitioners because many people have some experience in training CNNs. Once one has trained a baseline, one has a teacher available to obtain soft labels for free. Furthermore, it is anecdotally reported that training with soft labels outperforms training with hard labels \cite{hinton2015distilling}; thus, we feel it is natural to adopt teacher-student training in this work. We employ teacher-student learning with soft labels for our proposed prototype-based network to transfer the knowledge from a standard teacher CNN to a prototype-based student network to retain the predictive performance of the teacher. %consisting of the same CNN encoder architecture as the teacher and a prototype network component. 
In this section, we first introduce our prototype network architectures, the iterative prototype replacement algorithm, and the learning objectives. We also present the explainability methods for the network and the formulation of the outlier score with respect to the student architectures.
\begin{figure}[!htb]
    \centering
  \includegraphics[height=1.5in, width=0.8\linewidth]{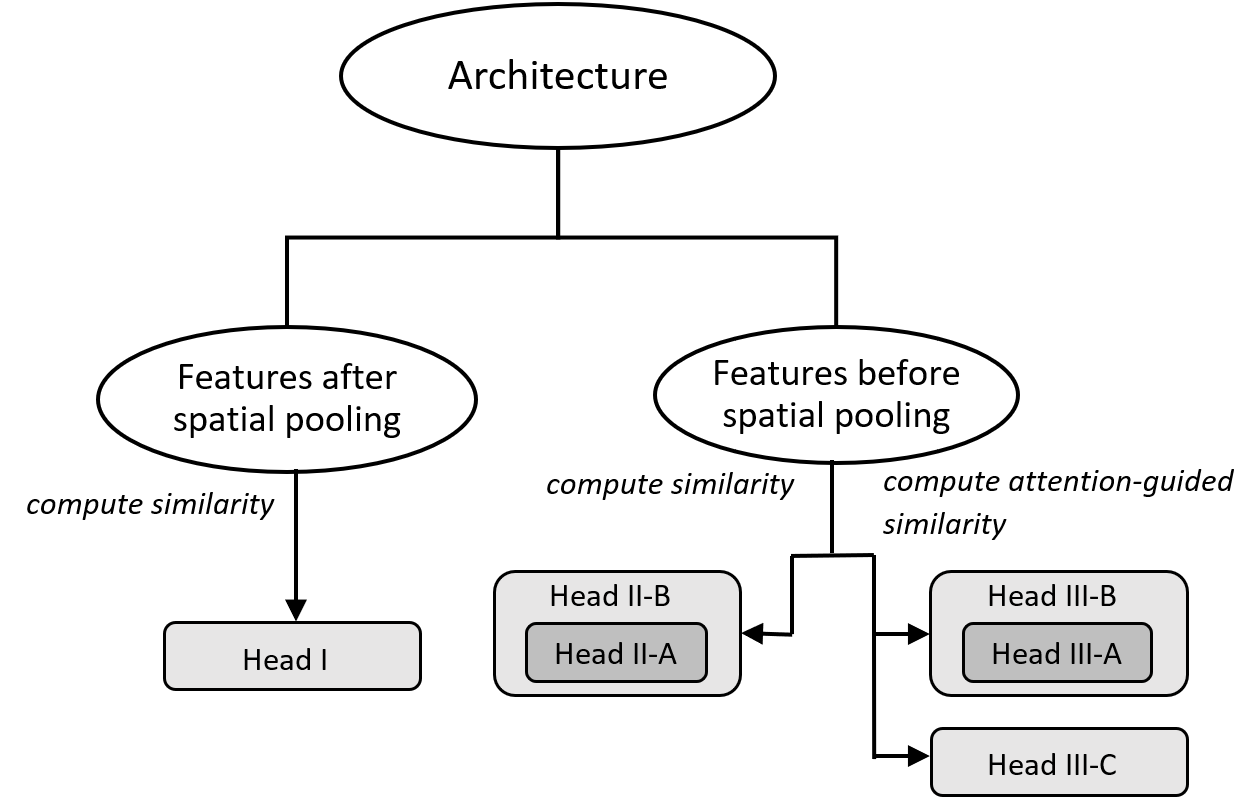}
  \caption{Taxanomy of the different categories of head architectures proposed for the student network.
}\label{taxanomy}
\end{figure}
%\subsection{Relation to Kernel SVM }\label{Relation to Kernel SVM}

Our proposed architecture for the student network is inspired by the dual formulation of SVMs \cite{cortes1995support} that solves the optimization problem 
$
\max_{\alpha_i \geq 0} \sum^{N}_{i}\alpha_i - \frac{1}{2}\sum^{N}_{j} \sum^{N}_{k} \alpha_j \alpha_k y_j y_k K(\vec{x_j}, \vec{x_k}),
$
subject to the constraints $0\leq\alpha_i\leq C, \ \forall i$ and $\sum^{N}_{i} \alpha_i y_i=0$. 
For a binary problem, the dual classifier takes the form of $f(\vec{x})=\sum^{N}_{i}\alpha_i y_i K(\vec{x_i},\vec{x})+ b$,
where $K(\vec{x_i},\vec{x})$ is a positive definite kernel. The solution is based on the set of similarities $\{K(\vec{x_i},\vec{x}) \ | \  1 \leq i \leq N \}$  between the training samples and the test sample $\vec{x}$ which can be used to compute statistics to quantify outliers \cite{10.1007/978-3-642-13062-5_16}. We employ architectures based on attention-weighted similarities beyond Mercer-Kernels. %We show later in Section \ref{Network Architecture} the relation of our method to the dual kernel SVM.
%%%%%%%%
\begin{figure*}[!htb]
\minipage{0.30\textwidth}
  \includegraphics[height=3.0in, width=\linewidth]{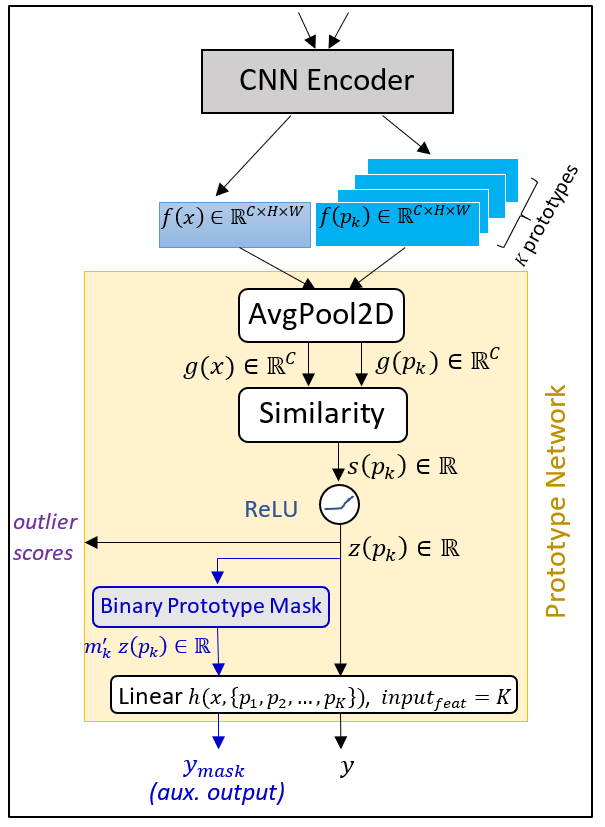}
  \caption{Head I : Similarity using features after spatial pooling. The $y_{mask}$ output branch  is the auxiliary output branch for iterative prototype replacement.
}\label{head1}
\endminipage\hfill
\minipage{0.30\textwidth}
  \includegraphics[height=3in, width=\linewidth]{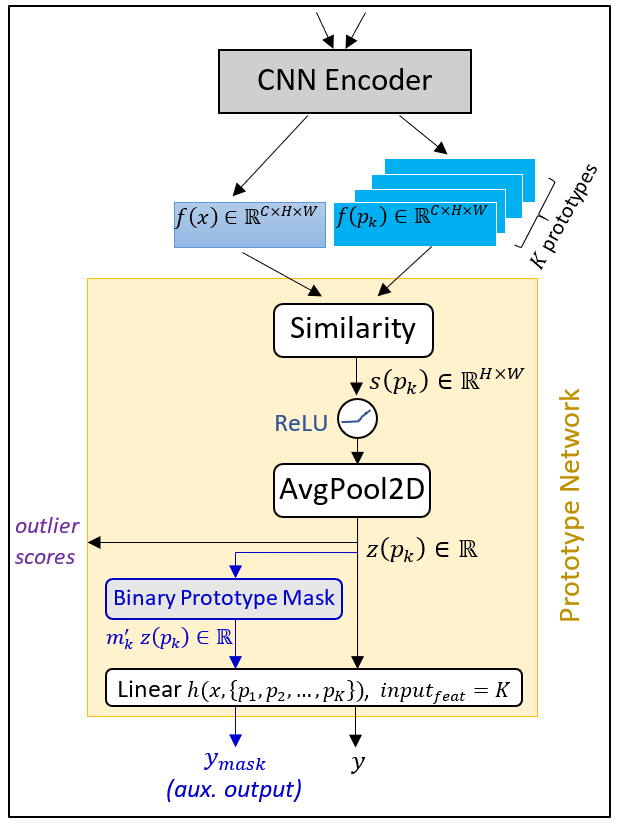}
  \caption{Head II : Similarity using features before spatial pooling. The $y_{mask}$ output branch is the auxiliary output branch for iterative prototype replacement.
}\label{head2}
\endminipage\hfill
\minipage{0.33\textwidth}%
\begin{center}
    
  \includegraphics[height=2.961in, width=0.95\textwidth]{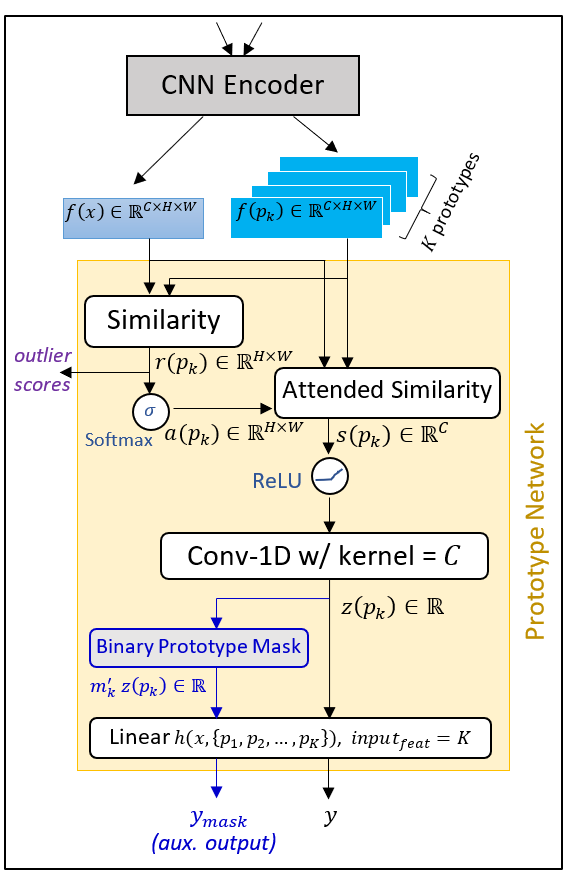}
  \end{center}
  \caption{Head III : Similarity w/attention using features before spatial pooling.  The $y_{mask}$ output branch  is the auxiliary output branch for iterative prototype replacement.
}\label{head3}
\endminipage
\end{figure*}
\subsection{Proposed Architecture for the Student Network}\label{Network Architecture}
%%%%%%%%%%%%%%%%%%%%%%%%%%%%%%%%%%%%%%%%%%%%
For the purpose of  providing meaningful explanations and detecting outliers, we introduce three generic head architectures\footnote{The term head architecture refers to the architecture used in the prototype network which is the portion of the network located after the CNN encoder. Figures \ref{head1}-\ref{head3} illustrate this.} for our prototype student network. Each architecture is evaluated for its performance in providing explanation and detecting outliers in Section \ref{Results}. For simplicity, we assume in this section that the prototypical examples are already selected based on our iterative prototype replacement algorithm. We will introduce the algorithm in Section \ref{Prototype Selection} later.

Let $f(\cdot\ ;\theta_1):\mathbb{R}^{C_0\times H_0 \times W_0} \rightarrow \mathbb{R}^{C\times H \times W}$ be the CNN feature extractor  before the spatial pooling layer with the learnable parameters $\theta_1$.  
Let $e(\cdot\ ;\theta_2):\mathbb{R}^{C\times H \times W} \rightarrow \mathbb{R}^{\tilde{C}} $ be the prototype network with the learnable parameters $\theta_2$, and let $\tilde{C}$ be the number of classes in the classification task. The overall student network is represented by $e \ \circ f$. For simplicity, we omit the parameters $\theta_1$ and $\theta_2$ and denote the feature map representation of the input sample $x$ as $f(x)$ in this section.

To make a prediction, we have to solve the problem of computing a similarity between the feature map $f(x)$ for a test sample $x$ and the feature $f(p)$ for a prototype $p$. The feature maps $f_{c,h,w}(\cdot)$ have a channel dimension $c$ and spatial dimensions, usually the two $h,w$. We observe that three larger categories of similarities can be defined, which correspond to the head classes \textbf{I}, \textbf{II} and \textbf{III}. The \textbf{Head I} follows the simplest idea to compute a similarity without any spatial dimensions; thus, it pools over the spatial dimensions before computing an inner product over the channel dimension. The \textbf{Head II} class computes similarities that incorporate the spatial dimensions. The difference between the two heads in the Head II class is whether one matches each spatial coordinate $(h,w)$ in the test sample to the same coordinate in the prototype as in Head II-A or whether one searches for every coordinate in the test sample for the best matching coordinate in the prototype using a maximum-operation as in Head II-B. Finally, the \textbf{Head III} class also uses the spatial dimensions of feature maps but additionally learns an attention weight over the spatial coordinates of the features in the test sample. This serves to weight the spatial regions in the test samples in the final similarity measures, whereas the \textbf{Head II} class treats each spatial coordinate $(h,w)$ of the test sample feature map with equal weight.

% \changemarker{\st{We explore various head architectures for the prototype-based student network to determine their suitability for explanation and identifying outliers. We categorize the head architectures for the prototype network into three classes: similarity after spatial pooling, similarity before spatial pooling, and attention-guided similarity before spatial pooling.}} 
Figure \ref{taxanomy} shows the taxonomy of the various head architectures for the student network and Figures \ref{head1}-\ref{head3} are visualizations of the head architectures. Our task differs from the standard image retrieval task as our task uses prototypes from the training set that participate in the prediction and thus share quantifiable prediction reasoning as the test sample, whereas the image retrieval task involves an exhaustive nearest neighbor search over the entire database that does not participate in the prediction and thus does not guarantee similar prediction reasoning between the test sample and its nearest neighbor.

Our proposed head architectures do not require any modification to the CNN encoder compared to \cite{arik2020protoattend}, which requires additional linear layers for queries and keys that add to the overhead during training. 

%%%%%%%
The \textbf{Head I} architecture is based on a spatial average pooling of the feature maps $f(\cdot)$, resulting in a vector $g(\cdot)$ that consists of only $C$ feature channels. This is followed by a cosine similarity over the feature channels $g(\cdot)$, as seen in Figure \ref{head1}:
\begin{equation}
    s^{\text{(I)}}(p_k)=\frac{g(x)^\top g(p_k)}{\|g(x)\|  \|g(p_k)\|}.
\end{equation}
%where $g_{x_i}[c]$ and $g_{p_k}[c]$ are the features $c$ in the feature vectors $g(x_i)$ and $g(p_k)$, respectively. 
The similarity output $s^{\text{(I)}}(p_k)$ falls in the range of $[-1,1]$ where a higher value indicates higher similarity between the input sample $x$ and the prototype $p_k$. 
The linear classification layer  %$h(x,\{p_1,p_2,...,p_K\})= \sum^{K}_{k=1} w_k ReLU(s(x,p_k))+b$ 
$h(x,\{p_1,p_2,...,p_K\})= \sum^{K}_{k=1} w_k ReLU(s^{\text{(I)}}(p_k))+b$
then predicts the output logit $y$. For this head architecture, the learnable prototype network parameters $\theta_2$ denote the parameters of the linear layer $h(\cdot)$.
The linear layer $h(\cdot)$  in Head I is consistent with the formulation of the dual kernel SVM. %described in Section \ref{Relation to Kernel SVM}. %The prediction for the dual kernel SVM depends on the support vectors, while our formulation depends on the set of prototypes $\{p_1,p_2,...,p_K\}$. 
The linear layers $h(\cdot)$ in Heads II and III as shown in Figures \ref{head2} and \ref{head3} will have similar dependency on a set of prototypes $\{p_1,p_2,...,p_K\}$, but they no longer correspond to kernels. %For all head architectures, the role of the auxiliary output branch in guiding the learning of prototype importance weight $\mathcal{M}$ used for prototype replacement will be discussed in Section \ref{Prototype Selection}.
%%%prototype replacement alg
\begin{algorithm*}[!ht]
\caption{Iterative Prototype Replacement Algorithm}
\label{alg:protoselect}
\renewcommand{\thealgorithm}{}
\floatname{algorithm}{}
\begin{algorithmic}[1]
\State \textbf{Initialization}: Let $S=D\cup P$ be the original training set with $n(S)=n(D)+n(P)=N+K$. $D$ as the new training set and  $P=%P_1\cup ...\cup P_{\tilde{c}} \cup... \cup P_{\tilde{C}}=
\{p_1,...,p_k,...,p_K \}$ as the  set of all prototypes. %$P_{\tilde{c}}$ is the set of prototypes from class $\tilde{c}$. 
%$\forall \tilde{c}$, $L$ prototypes from class $\tilde{c}$ are randomly sampled from the original training set $S$  and assign as $P_{\tilde{c}}$.
The initial set $P$ consists of an equal number of samples per class that are randomly sampled from the original training set $S$.
%For each class, $L$ samples are randomly sampled as prototypes from the original training set $S$.
Let $\theta_1$ and $\theta_2$ be the parameters of the CNN feature extractor and the prototype network, respectively, and $\mathcal{M} \in \mathbb{R}^K$ be the prototype importance weight used for replacement.
\While{$e\leq$ number of epochs}

        \While{ $t \leq$ number of iterations}

                \State Perform a forward pass to obtain input features $f(x)$, set of prototype features $\{f(p_k) \}$, corresponding input features  \indent \ \ \  $\{z(p_k)\}$ of the linear classification layer $h(\cdot)$, and the output logits $y$. (For Head I, we compute $g(x)$ and $\{g(p_k) \}$  \indent \ \ \ instead of $f(x)$ and  $\{f(p_k) \}$.)
                
                \State Compute threshold $\tau_t$ as the $p$-th smallest weight in $\mathcal{M}$. 
             
                \State Compute binary prototype mask $\mathcal{M}'(\tau_t)$  from $\mathcal{M}$ as shown in Eq. \eqref{mask_eqn}.
                
                \State Perform a forward pass in $h(\cdot)$ using the masked input to obtain the auxiliary output $y_{mask}$ as shown in  Eq. \eqref{masked_output_eqn}. 
                \State Optimize parameters $\theta_1,\theta_2$ and $ \mathcal{M}$ based on the loss function \eqref{total_loss}.

            %% if needed later on if else for method replacement type %%%
            
        \EndWhile

    \State $P\gets P\setminus P_{replace} \cup D_{rand}$, where $P_{replace}$ is the set of $p$ prototypes with the smallest weight in $\mathcal{M}$, and $D_{rand}$ is the \hspace*{4mm} set of random samples comprising the replaced prototype classes sampled from $D$ to replace $P_{replace}$. 
    
    \State $D\gets D\setminus D_{rand} \cup P_{replace}$
    
    \State Reinitialize only the weights in $\mathcal{M}$ where prototypes are replaced.
\EndWhile
\end{algorithmic}\end{algorithm*}

The \textbf{Head II} architecture in Figure \ref{head2}, uses the features before spatial pooling to compute similarity scores. Therefore, the similarity layer computes $s^{\text{(II)}}(p_k) \in \mathbb{R}^{H\times W}$, i.e.,~one similarity score for each spatial position $(h,w)$. We explore two types of similarity functions for this head architecture: A) similarity between the input and prototype at the same spatial position $(h,w)$ and 
B) the maximum similarity between input and prototype at a given \emph{input spatial location} $(h,w)$.
%B) the maximum similarity between input and prototype over a set of spatial location pairs.
We refer to the former as Head II-A and the latter as Head II-B. The similarity operation for Head II-A (Eq.\eqref{Head II-A}) and Head II-B (Eq.\eqref{Head II-B}) at the spatial location $(h,w)$ is defined as:
\begin{align}
\label{Head II-A}
    %s^{\text{(II-A)}}_{h,w}(p_k)&=\sum_c \frac{f_{c,h,w}(x)}{n_{h,w}(x)}\cdot \frac{f_{c,h,w}(p_k)}{n_{h,w}(p_k)},\\
        s^{\text{(II-A)}}_{h,w}(p_k)&=\sum_c \hat{f}_{c,h,w}(x)\cdot \hat{f}_{c,h,w}(p_k),\\
\label{Head II-B}
    s^{\text{(II-B)}}_{h,w}(p_k)&=\max_{h',w'}  \sum_c \hat{f}_{c,h,w}(x)  \cdot  \hat{f}_{c,h',w'}(p_k),\\
%n_{h,w}(x) &= \sqrt{\sum_c [f_{c,h,w}(x)]^2} =  \|f_{\cdot,h,w}(x)\|_2,\\
%\hat{f}_{c,h,w}(\cdot)&=\frac{f_{c,h,w}(\cdot)}{n_{h,w}(\cdot)},\\
\hat{f}_{c,h,w}(\cdot)&=\frac{f_{c,h,w}(\cdot)}{\|f_{\cdot,h,w}(\cdot)\|_2} = \frac{f_{c,h,w}(\cdot)}{\sqrt{\sum_c [f_{c,h,w}(\cdot)]^2}},
\end{align}
where $f_{c,h,w}(x)$ and $f_{c,h,w}(p_k)$ are the $c$-th features at the spatial location $(h,w)$ from the feature maps $f(x)$ and $f(p_k)$, respectively. Head II-A compares the pairwise similarity of the spatial features between $f(x)$ and $f(p_k)$ along the same spatial location, whereas Head II-B considers all spatial locations of $f(p_k)$ to find the location that is most similar to the specified spatial location in  $f(x)$. Therefore, Eq. \eqref{Head II-B} ensures that the similarity at the spatial location $(h,w)$ is computed only between the location $(h,w)$ in  $x$ and the most relevant location $(h',w')$ in $p_k$, i.e., the pair that locally gives the highest cosine similarity score. Its similarity operation is spatially invariant, as it considers all spatial locations of prototype $p_k$ in identifying the most similar feature to the input $x$.
Head II-A (Eq. \eqref{Head II-A}) can be viewed as a special case of Head II-B (Eq. \eqref{Head II-B}). The operation of Head II-B reduces to Head II-A when the maximum similarity between input and prototype for a given \emph{input spatial location} $(h,w)$, also occurs at the prototype location $(h,w)$. Thus, the Head II-B architecture is considered the general case of Head II-A. 
For this category of architectures, the learnable prototype network parameters $\theta_2$ denote the parameters of the linear layer $h(\cdot)$.
%In the remaining parts of the paper, $\frac{f_{c,h,w}(\cdot)}{n_{h,w}(\cdot)}$ will be denoted as $\hat{f}_{c,h,w}(\cdot)$. 
%we remove this because we remove Head III-C and Head III-D from the architecture. Consider putting this as ablation studies.
Note that Head II-B, as a maximum over inner products, no longer defines a kernel, as the maximum of two positive definite matrices is not guaranteed to be a positive definite. Thus, we employ a nonkernel-based similarity.

The \textbf{Head III} architecture employs a common attention mechanism to compute weighted similarities. For the Head III architecture as shown in Figure \ref{head3}, the similarity scores from Eq. \eqref{Head II-A} and \eqref{Head II-B} are passed into a softmax $\sigma(\cdot)$ to obtain the following attention maps:
%For Head III architecture as shown in Figure \ref{head3}, the similarity scores from Eq. \eqref{Head II-A} and \eqref{Head II-B} are passed into a softmax $\sigma(\cdot)$ to obtain the attention map $a(x,p_k) \in \mathbb{R}^{ H \times W}$, i.e., $a(x,p_k)=\sigma( r(x,p_k))$. For Head III-A $r_{h,w}(x,p_k)$ is the similarity score $ s_{h,w}(x,p_k)$ from Eq. \eqref{Head II-A}, and for Head III-B $r_{h,w}(x,p_k)$ is $ s_{h,w}(x,p_k)$ from Eq. \eqref{Head II-B}. The attention map $a(x,p_k)$ will be used together with the unnormalized features $f(x)$ and $f(p_k)$ to compute a spatially weighted similarity vector $s(x,p_k) \in \mathbb{R}^C$. The $s(x,p_k)$ operation for the channel $c$ in the attended similarity layer  for Head III-A (Eq. \eqref{Head III-A}) and Head III-B (Eq. \eqref{Head III-B}) is defined as:
\begin{align}
a^{\text{(III-A)}}_{h,w}(p_k) &= \sigma(s^{\text{(II-A)}}_{h,w}(p_k)),\\
\label{attention_headIII-B}
a^{\text{(III-B)}}_{h,w}(p_k) &= \sigma(s^{\text{(II-B)}}_{h,w}(p_k)).
\end{align}
% \changemarker{The softmax operation normalizes the scores in the range of $[-1,1]$ to $[0,1]$ where a higher}
Head III-A and Head III-B are the attention-augmented counterparts of Head II-A and Head II-B, respectively:
\begin{align}
%a^{\text{(III-A)}}_{h,w}(p_k) &= \sigma(s^{\text{(II-A)}}_{h,w}(p_k)) \\
\label{Head III-A}
    s^{\text{(III-A)}}_{c}(p_k)&=\sum_{h,w} a^{\text{(III-A)}}_{h,w}(p_k)\cdot f_{c,h,w}(x)\cdot f_{c,h,w}(p_k),
    \end{align}
\begin{align}
%a^{\text{(III-B)}}_{h,w}(p_k) &= \sigma(s^{\text{(II-B)}}_{h,w}(p_k)) \\
    \label{Head III-B} s^{\text{(III-B)}}_{c}(p_k)&=\sum_{h,w} a^{\text{(III-B)}}_{h,w}(p_k)\cdot f_{c,h,w}(x)\cdot f_{c,h_{\max},w_{\max}}(p_k).
\end{align}
The indices $h_{\max}$ and $w_{\max}$ in Eq. \eqref{Head III-B} are the selected $h'$ and $w'$ indices used in the computation of $s^{\text{(II-B)}}_{h,w}(p_k)$.  
The attention map $a(p_k) \in \mathbb{R}^{H\times W}$ is applied over the spatial features $f(x) \in \mathbb{R}^{C\times H \times W}$ and $f(p_k) \in \mathbb{R}^{C\times H \times W}$ to give more weight to the important regions (i.e., spatial regions between $f(x)$ and $f(p_k)$ with high similarity) for the neural network to learn and focus on these regions.
The output from the attended similarity layer which represents a weighted spatial similarity for every channel, 
is passed into a convolutional-1D layer with kernel size $C$ to compute a weighted sum of the $C$ features. Since $C \gg H\times W$, we avoid using simple average spatial pooling over the $C$ features to obtain better explanation heatmaps.
The convolutional-1D layer has a minimum weight clipping at zero to ensure the outputs are always nonnegative in order to apply the explainability method described in Section \ref{XAI}. 
Additionally, the use of individual attention maps for the respective features $f(x)$ and $f(p_k)$ is also investigated in Head III-C:
\begin{align}
a^{\text{(III-C)}}_{h,w}(p_k) &= \sigma \left(\max_{u,v}\sum_c \hat{f}_{c,u,v}(x) \cdot \hat{f}_{c,h,w}(p_k)\right),\label{attn_fpk}\\
\label{Head III-C}
    s^{\text{(III-C)}}_{c}(p_k)&=\sum_{h,w} a^{\text{(III-B)}}_{h,w}(p_k)\cdot f_{c,h,w}(x)  \cdot a^{\text{(III-C)}}_{h,w}(p_k)\cdot f_{c,h,w}(p_k),
\end{align}
where $a^{\text{(III-B)}}_{h,w}$ is the attention score from Eq. \eqref{attention_headIII-B} for $f(x)$ at the \emph{input spatial location} $(h,w)$, whereas $a^{\text{(III-C)}}_{h,w}$ is the attention score for $f(p_k)$ at the \emph{prototype spatial location} $(h,w)$, as can be seen by the computation of $\max_{u,v}(\cdot)$  over the spatial locations of $\hat{f}(x)$. %The operation $a'_{h,w}(x,p_k)=\sigma( r'(x,p_k))_{h,w}$ at the spatial location $(h,w)$ is the softmax function $\sigma(\cdot)$ over the maximum dot product between input $x$ and prototype $p_k$ at a specified \emph{prototype location} $(h,w)$, i.e.,  $r'_{h,w}(x,p_k)=\max_{h',w'}(\sum_c \hat{f}_{c,h',w'}(x) \cdot \hat{f}_{c,h,w}(p_k))$. 
Eq. \eqref{attn_fpk} computes an attention map by considering all spatial locations of $f(x)$ to find the most similar location to the specified location in $f(p_k)$ as opposed to the attention map based on Eq. \eqref{Head II-B}. The attention map $a^{\text{(III-B)}}(p_k) \in \mathbb{R}^{H\times W} $ with weights denoting the importance of the spatial positions in $f(x)$ based on its similarity with $f(p_k)$,
is applied over the feature $f(x)$. On the other hand, the attention map $a^{\text{(III-C)}}(p_k) \in \mathbb{R}^{H\times W}$ with importance weights for the spatial positions in $f(p_k)$ based on its similarity with $f(x)$, is applied over the feature $f(p_k)$.
For the head architectures in this category, the learnable prototype network parameters $\theta_2$ are the parameters of the convolutional-1D and the linear classification $h(\cdot)$ layers.
%%% uncomment this for Head III-C and Head III-D architectures. Otherwise put as ablation study
% Finally, the use of different attention maps for features $f(x_i)$ and $f(p_k)$ is also investigated. The following defines the operation for Head III-C (Eq. \eqref{Head III-C}) and Head III-D  (Eq. \eqref{Head III-D}).
% \begin{align}\label{Head III-C}
%     s_{c}(x_i,p_k)&=\sum_{h,w} (a_{h,w}(x_i,p_k)\cdot f_{x_i}[c,h,w])\cdot (a'_{h,w}(x_i,p_k)\cdot f_{p_k}[c,h,w]),\\
% \label{Head III-D}
%     s_{h,w}(x_i,p_k)&=\sum_{c} (a_{h,w}(x_i,p_k)\cdot f_{x_i}[c,h,w])\cdot (a'_{h,w}(x_i,p_k)\cdot f_{p_k}[c,h,w]),
% \end{align}
% where $a'_{h,w}(x_i,p_k)=softmax( r'(x_i,p_k))_{h,w}$  is the attention score at spatial location $(h,w)$ and  $r'_{h,w}(x_i,p_k)=\max_{h',w'}(\sum_c \hat{f}_{x_i}[c,h',w'] \cdot \hat{f}_{p_k}[c,h,w])$. The attended similarity layer for Head III-C sums over the spatial dimension $H,W$, while  Head III-D sums over the feature dimension $C$.

\subsection{Auxiliary Output Branch for Iterative Prototype Replacement}\label{Prototype Selection}
For the selection of representative prototypes, one can directly learn an importance mask over the entire training set %together with the neural network parameters ($\theta_1$ and $\theta_2$) 
and select those samples with high importance as prototypes. This approach, however, does not scale well to large sample sizes as the number of parameters that must be learned increases proportionally. For approaches such as \cite{arik2020protoattend}, the database of prototype candidates used must be sufficiently large for successful training as reported in their paper.
Instead of learning an importance mask over the entire training set or a large candidate database, we propose to learn a smaller importance mask over a fixed number of $K$ prototypes. We note that $K$ is very much smaller than the size of the candidate database \cite{arik2020protoattend} used during training or inference. Our approach scales well to large sample sizes since the number of parameters to be learned can now be constrained by $K$ which also corresponds to the final number of prototypes used during inference.
We introduce an auxiliary output branch to guide the network to learn a prototype replacement scheme that replaces uninformative prototypes (out of the $K$ prototypes) with new samples based on the importance or contribution of the current prototypes to the decision boundary of the network. 
Prototypes that are assigned larger importance weights have higher importance and are unlikely to be replaced. On the other hand, prototypes with smaller importance weights are likely to be replaced with other samples from the training set due to their minimal contribution to the decision boundary of the network.

During training, the network learns a prototype importance weight $\mathcal{M} \in \mathbb{R}^K$ via an auxiliary loss to quantify the importance of the $K$ prototypes based on their contribution to the prediction. 
The auxiliary loss imposes a penalty when informative prototypes that influence the decision boundary of the network are removed, i.e.,  masked out. 
The network parameters $\theta_1$ and $\theta_2$ and the prototype importance weight $\mathcal{M}$ are jointly optimized. The formulation of the auxiliary loss will be elaborated further in 
Section \ref{Learning Objectives}.

The full algorithm for the iterative prototype replacement method is presented in Algorithm \ref{alg:protoselect}. %As stated in the algorithm, the binary prototype mask $\mathcal{M'}$ is computed from the prototype importance weight $\mathcal{M}$.
%We define the binary prototype mask $\mathcal{M'}$  used for prototype replacement as:
We compute the binary prototype mask $\mathcal{M}'(\tau_t)$ from the prototype importance weight $\mathcal{M}$ using the threshold $\tau_t$ at iteration $t$ as:
\begin{align}\label{mask_eqn}
   \mathcal{M}'(\tau_t)= \frac{\max \  (0,\mathcal{M} - \tau_t )}{|\mathcal{M}-\tau_t|} \in \{0,1\}. 
\end{align}
%where $\tau_t$ is the assigned threshold at iteration $t$. 
We use continuous ReLU activation with normalization as inspired by the hard shrinkage operation in \cite{gong2019memorizing} to avoid issues with the backpropagation of a discontinuous 0-1 step function.
%The computed binary prototype mask $\mathcal{M'}$ consists of values $0$ and $1$. 
%There are many ways to assign the threshold $\tau_t$. 
For simplicity, we assign $\tau_t$ as the $p$-th smallest weight in $\mathcal{M}$ to zero out a fixed number of $p$ prototypes (in the input of the linear classification layer $h(\cdot)$) with the smallest weight in $\mathcal{M}$.
To compute the auxiliary output $y_{mask}$, we multiply the binary prototype mask $\mathcal{M}'(\tau_t)$ elementwise with the features $z(p_k) \in \mathbb{R}$ at the input of the final linear classification layer $h(\cdot)$ corresponding to each prototype as follows:% over the input of the linear classification layer $h(\cdot)$ and perform a forward pass in the layer to obtain $y_{mask}$ as follows: %The following defines the forward operation in $h(\cdot)$ to obtain the auxiliary output $y_{mask}$:
\begin{align}
    \label{masked_output_eqn} y_{mask} =h(x,\{p_1,p_2,...,p_K\})= \sum^{K}_{k=1} w_k m'_k  z(p_k)+b , 
\end{align}
where $m'_k \in \{0,1\}$ is the weight in  $\mathcal{M}'(\tau_t)$, and $w_k \in \mathbb{R}$ and $b \in \mathbb{R}$ are the parameters of the linear layer $h(\cdot)$.
%where $z(x,p_k) \in \mathbb{R}$ is the input feature of the linear layer $h(\cdot)$, $m'_k \in \mathbb{R}$ is the weight in  $\mathcal{M}'$, and $w_k \in \mathbb{R}$ and $b \in \mathbb{R}$ are the parameters of the layer $h(\cdot)$. %The auxiliary output $y_{mask}$ is used in the auxiliary loss. 
At the end of each epoch, we replace the $p$ prototypes that have the smallest weight in $\mathcal{M}$ with new samples %comprising of the replaced classes
from the training set $D$. The new set of prototypes $P$ also maintains an equal number of samples per class during training.

The prototype importance weight $\mathcal{M}$ can also be used to prune prototypes and relevant neurons. This will be discussed in Section \ref{Pruning of Prototypes and Relevant Neurons}.

We remark that it is not of interest to iterate through all the training samples when choosing optimal prototypes, as this will not scale well to large datasets. Our algorithm is based on the idea that one can find prototypes that may be similar to a number of training samples and thus can avoid iterating through all training samples when selecting prototypes. In future work, we will consider presorting the prototypes based on their similarity to maintain an order over training samples during learning to speed up model convergence. The prototype-based Head III architectures with higher architectural complexities than Head I and II architectures have greater training overhead than a standard CNN teacher network due to the computation of attention maps  and training time which increases proportionally with the number of prototypes used. Additional details are provided in Section S-I of the Supplementary Material.

\subsection{Learning Objective}\label{Learning Objectives}
%Let $(x,\tilde{y})$ be an input-label sample from the training set $D$ and let $P=\{(p_k, \tilde{q}_k) \ | \ 1\leq k \leq K \}$ be the set of prototypes.
We define the set of input training samples with their ground truth to be $D=\{(x_i,\tilde{y}_i) \ |\ 1\leq i \leq N \}$ and the set of prototypes with their class labels to be $P=\{(p_k, \tilde{q}_k) \ | \ 1\leq k \leq K \}$. %The notations  $\tilde{y}_i$ and $\tilde{q}_k$ denote the ground truth class labels for the input sample $x_i$ and prototype $p_k$, respectively. 
For an input sample $x$, we denote the output logits of the teacher network as $y_T$ and the logits and auxiliary output of the student network as $y$ and $y_{mask}$, respectively. The predicted class label from the student network for the input sample $x$ is represented by $y_{pred}=\arg\max y$. 
The following defines the objective function for our student network: 
\begin{multline}\label{total_loss}
\min_{\theta_1, \theta_2, \mathcal{M}} \ \mathcal{L}(\tilde{y}, \sigma(y)) + \lambda_1\mathcal{L}(\sigma(y_{T}),   \sigma(y)) \\ + \lambda_2\mathcal{L}(y_{pred},   \sigma(y_{mask})) +
\lambda_3\mathcal{J}\left((x,\tilde{y}), (p, \tilde{q})\right),
\end{multline}
% \begin{equation} \label{total_loss}
% \mathcal{L}_{final} = \mathcal{L}_{CE} + \lambda_1\mathcal{L}_{CE\_Soft} + \lambda_2\mathcal{L}_{Dist},
% \end{equation}
where $\mathcal{L}(\cdot)$ is the cross-entropy loss, $\mathcal{J}(\cdot)$ is a loss term based on the squared  $L_2$-norm, and $\sigma(\cdot)$ is the softmax. 
%The first two terms are the standard classification and distillation losses commonly employed in the training of a student network. 

The first two terms are the losses commonly employed in the training of a student network. The second term, also known as the distillation loss, is the cross-entropy loss with the teacher soft labels as ground truth to encourage the student network to learn predictions resembling the teacher network.

%The third term $\mathcal{L}(y_{pred},   \sigma(y_{mask}))$ encourages the learning of parameters such that the network class prediction remains unchanged while relying only on a subset of prototypes which leads to the dichotomization of important and less important prototypes.
The third term $\mathcal{L}(y_{pred},   \sigma(y_{mask}))$ is used for the learning of prototype importance weight $\mathcal{M} \in \mathbb{R}^K$. We use the standard cross-entropy loss as the auxiliary  loss but with the predicted class label and the probability $\sigma(y_{mask})$ as the input argument to the standard cross-entropy loss rather than using the ground truth class $\tilde{y}$ and the probability output $\sigma(y)$ from the main network branch.
The auxiliary loss enables the network to learn  prototype importance weight $\mathcal{M} \in \mathbb{R}^K$ which gives larger weights to prototypes that will cause a large change in the decision boundary of the network when removed, and lower importance weights to  prototypes that cause only a small change to the decision boundary if removed. Our proposed solution is based on the idea of removing uninformative prototypes with minimal disruption to the decision boundary using the predicted class label $y_{pred}$ instead of the ground truth class label $\tilde{y}$ to guide the learning of parameters $\mathcal{M} \in \mathbb{R}^K$ since
%(i.e., prototypes that cause only minimal change in the decision boundary when removed) 
we believe the network can learn better by replacing  uninformative prototypes with other potentially more informative prototypes. This allows the replacement of lower ranked prototypes
that have a lower impact on the prediction accuracy in the prototype replacement step.

%The third term $\mathcal{L}(y_{pred},   \sigma(y_{mask}))$ is introduced to encourage the replacement of uninformative or less important prototypes, i.e., prototypes that do not contribute or contribute minimally to the decision boundary of the classifier as will be shown in its definition further below. The network learns a scheme to replace prototypes which cause the least change in the network prediction upon their removal in order to ensure minimal distortion of the decision boundary of the network. To achieve this, we use the cross-entropy loss as the auxiliary loss with predicted labels $y_{pred}$ as the ground truth and the auxiliary output $y_{mask}$ to learn the prototype importance weight $\mathcal{M}$. The auxiliary loss enforces a penalty when informative prototypes are masked out. Thus, the prototype importance weight $\mathcal{M}$ has a correlation with the importance of each prototype in the construction of the network prediction.
The fourth loss term $\mathcal{J}\left((x,\tilde{y}), (p, \tilde{q})\right)$ encourages the feature space of the input sample $x$ to be close to the set of prototypes from the same class but farther away from the other classes. The formulation of this objective term varies slightly between the different head architectures depending on their similarity operation.

For the Head I architecture, the objective term  $\mathcal{J}\left((x,\tilde{y}), (p, \tilde{q})\right)$ is defined as:
\begin{equation*} \label{head1-distloss}
\frac{1}{|D||P|} \sum_{i=1}^{|D|} \sum_{k=1}^{|P|} ( \ \| \hat{g}(x_i)-\hat{g}(p_k) \|_2^2 \ )^{\alpha},
% where \ \alpha=1, if \ \tilde{y}=\tilde{q}, \ otherwise \ \alpha=-1.
\end{equation*}
\begin{equation}
\text{where} \ \alpha=1, \ \text{if} \ \tilde{y}_i=\tilde{q}_k, \ \text{else} \ \alpha=-1.
\end{equation}
%%%%%%
% \begin{equation} \label{head1-distloss}
%  \frac{1}{NK} \sum^N_{i=1} \sum^K_{k=1} ( \ \| \hat{g}(x_i)-\hat{g}(p_k) \|^2 \ )^{\alpha_{i,k}} \ ,
% \end{equation}
% where $\alpha_{i,k}=1$, if $y_i=\tilde{y}_k$ (same class), otherwise $\alpha_{i,k}=-1$ (different class). 
For the Head II-A and Head III-A architectures, the objective function $\mathcal{J} \left((x,\tilde{y}), (p, \tilde{q})\right)$ is defined as: 
\begin{equation*}
\frac{1}{|D||P|} \sum_{i=1}^{|D|} \sum_{k=1}^{|P|} \bigg[\frac{1}{HW} \sum^{H,W,C}_{h,w,c}(\hat{f}_{c,h,w}(x_i)-\hat{f}_{c,h,w, }(p_k))^2\bigg]^{\alpha},
\end{equation*} 
\begin{equation}\label{Head II-A and Head III-A}
\text{where} \ \alpha=1, \ \text{if} \ \tilde{y}_i=\tilde{q}_k, \ \text{else} \ \alpha=-1.
\end{equation}
% \begin{equation} \label{Head II-A and Head III-A}
% \mathcal{L}_{Dist} = \frac{1}{NK} \sum^N_{i=1} \sum^K_{k=1} \left[ \ \frac{1}{HW} \sum^{H,W}_{h=1,w=1} \sum_{c=1}^C (\hat{f}_{x_i}[c,h,w]-\hat{f}_{p_k}[c,h,w])^2 \ \right]^{\alpha_{i,k}}.
% \end{equation}
The term inside the square bracket is simply the average squared $L_2$-norm computed for the channel dimension over all spatial positions $(h,w)$.
Similarly, the objective function $\mathcal{J} \left((x,\tilde{y}), (p, \tilde{q})\right)$ for  the Head II-B and Head III-B architectures is defined as:
\begin{equation*} 
\frac{1}{|D||P|}  \sum_{i=1}^{|D|} \sum_{k=1}^{|P|}\bigg[\frac{1}{HW} \sum^{\mathclap{H,W,C}}_{\mathclap{h,w,c}} (\hat{f}_{c,h,w}(x_i) -\hat{f}_{c,h_{\max},w_{\max}}(p_k))^2 \bigg]^{\alpha} ,
\end{equation*}
\begin{equation}\label{Head II-B and Head III-B}
\text{where} \ \alpha=1, \ \text{if} \ \tilde{y}_i=\tilde{q}_k, \ \text{else} \ \alpha=-1.
\end{equation}
% \begin{equation} \label{Head II-B and Head III-B}
% \mathcal{L}_{Dist} = \frac{1}{NK}  \sum^{N,K}_{i,k} \left[ \ \frac{1}{HW} \sum^{H,W,C}_{h,w,c}  (\hat{f}_{x_i}[c,h,w]-\hat{f}_{p_k}[c,h_{\max},w_{\max}])^2 \ \right]^{\alpha_{i,k}} ,
% \end{equation}
%%uncomment below if introduce Head III-C and Head III-D
% and $\mathcal{L}_{Dist}$ for Head III-C and Head III-D is expressed as $\mathcal{L}_{Dist}=\mathcal{L}_{Dist_{x}}+ \mathcal{L}_{Dist_{p}}$, where $\mathcal{L}_{Dist_{x}}$ is from Eq. \eqref{Head II-B and Head III-B} and $\mathcal{L}_{Dist_{p}}$ can be defined in a similar manner.
The different formulation for Eq. \eqref{Head II-A and Head III-A} and Eq. \eqref{Head II-B and Head III-B} is due to the two different types of similarity operations introduced in the head architectures, as discussed in Section \ref{Network Architecture}.
For Head III-C architecture, the objective function $\mathcal{J}(\cdot)$ is expressed as 
\begin{equation}
\mathcal{J} \left((x,\tilde{y}), (p, \tilde{q})\right) = \mathcal{J}_x \left((x,\tilde{y}), (p, \tilde{q})\right) + \mathcal{J}_p \left((x,\tilde{y}), (p, \tilde{q})\right),
\end{equation}
where the first term is equivalent to Eq. \eqref{Head II-B and Head III-B} and the second term is defined accordingly based on the $a^{\text{(III-C)}}_{h,w}(p_k)$ operation for $f(p_k)$ in Head III-C, i.e., use $\hat{f}_{c,h_{max},w_{max}}(x_i)$ instead of $\hat{f}_{c,h,w}(x_i)$ and vice versa for $p_k$ in the loss $\mathcal{J}_p(\cdot)$.

\subsection{Explanation for Top-$k$ Nearest Prototypical Examples}\label{XAI}
The student network provides a prediction explanation in the form of:
\begin{enumerate*}
\item a prediction explanation with top-$k$ nearest prototypical examples and 
\item pixel evidence of similarity between the prediction sample and the prototypical examples.
\end{enumerate*}
We define a prototype similarity score $u_k$ that ranges between $[0,1]$ to identify the top-$k$ nearest prototypical examples for a given input sample. %A large prototype similarity score represents a small distance between the prototype and the given input sample. 
We elaborate on the formulation of the prototype similarity score $u_k$ in Section \ref{outlier} as the scores are also used to formulate the outlier score. In the following paragraph, we recapitulate the explanation method to compute pixel evidence of similarity between each pair of input-prototype.

To show evidence of similarity between the input sample and the prototypes in the pixel space, we use a posthoc explainability method, the Layer-wise Relevance Propagation (LRP) algorithm \cite{montavon2019layer,bach2015pixel} to compute a pair of LRP heatmaps for each pair of input-prototype. %The LRP algorithm backpropagates the prediction score $h(x)$ to the input space of $x$ to compute LRP heatmap for explanation. 
For a pair of input-prototype, their respective  heatmaps comprise positive and negative  relevance scores (also known as evidence) in the pixel space. The relevance scores represent the pixels' contribution to the prediction score.
%The algorithm is based on the conservation principle 
%$
%\sum_{d} R_d^{(l)} = \sum_{j} R_j^{(l+1)}
%$,
%where $R_d^{(l)}$ is the relevance score associated with neuron $d$ in layer $l$.
%%%%%%
% The conservation principle guarantees that the sum of the pixel scores of the LRP heatmap, $\sum_{d}h_d$ would equal to the prediction score, $h(x)$, i.e., $\sum_{d}h_d=h(x)$ where $h_d=R_d^{(1)}$.
%For the input layer, we used the following $z^B$-rule:
%$
%R_d^{(l)}= \sum_j \frac{x_dw_{d,j}-l_dw_{dj}^+ - h_d w_{dj}^-}{\sum_d x_dw_{d,j}-l_dw_{dj}^+ - h_d w_{dj}^-} \ R_j^{(l+1)}.
%$
For the 2D-convolutional layers, we used the following LRP-$\alpha\beta$ rule to compute  relevance $R_d^{(l)}$  for the neuron $d$ at the current layer $l$:
\begin{equation}
R_d^{(l)}=\sum_j \left( \ \alpha \ \frac{(a_dw_{dj})^+}{\sum_{0,d}(a_d w_{dj})^+} -\beta \ \frac{(a_dw_{dj})^-}{\sum_{0,d}(a_d w_{dj})^-} \ \right) R_j^{(l+1)},  
\end{equation}
where $a_d$ is the lower-layer activation from the input neuron $d$ at layer $l$, $w_{dj}$ is the weight connecting the neuron $d$ at layer $l$ to the higher-layer neuron $j$ at the succeeding layer $l+1$, $R_j^{(l+1)}$ is the relevance for the neuron $j$ at the succeeding layer $l+1$, $\alpha-\beta=1$, $\alpha >0$, and $\beta \geq 0$.
For the other layers, including the similarity layer (Heads I and II) and the attended similarity layer (Head III), we use the following  LRP-$\epsilon$ rule: 
\begin{equation}
R_d^{(l)}=\sum_j \frac{a_dw_{dj}}{\epsilon+\sum_{0,d} a_d w_{dj}} \ R_j^{(l+1)}.
\end{equation}
The similarity/attended similarity layer is treated as a linear layer, which enables the easy application of the LRP-$\epsilon$ rule  in the standard manner without resorting to more complex LRP methods such as  BiLRP \cite{eberle2020building}. For Head III architectures, the relevance scores flow through the attended similarity layer branch only while treating the attention weights from the similarity layer as constants. 

From the last linear layer up to the similarity/attended similarity layer $l$, we compute the relevance scores $R_{(\cdot)}^{(l)}(x,p_k)$ for each pair of $(x, p_k)$. %In classification problems, LRP is usually computed by initializing selected outputs of the logit layer with relevance scores. 
%Here, we initialize the outputs of the similarity layer instead. 
Using  $R_{(\cdot)}^{(l)}(x, p_k)$ at the similarity/attended similarity layer $l$, we compute the LRP heatmap for $x$ by backpropagating the relevances to the input space of $x$ through the CNN encoder. The heatmap for $p_k$ is also computed in a similar manner (for Head II-B and Head III-B architectures with $\max_{h',w'}(\cdot)$ operation, the relevance scores are accumulated at the relevant $h'$ and $w'$ indices before backpropagating to the input space of $p_k$).
%To compute the LRP heatmap for $x$ based on the similarity  between $x$ and $p_k$, the relevance score $R_{(\cdot)}^{(l)}(x, p_k)$ at the similarity/attended similarity layer $l$ is backpropagated to the input space of $x$ and heatmap for $p_k$ is also computed in a similar manner (for architectures with $\max(\cdot)$ similarity operation, relevance scores are accumulated at $h_{max}$ and $w_{max}$ indices before backpropagating to the input space of $p_k$).
%For the layers between the input $x$ and the similarity/attended similarity layer, the relevance map for $x$ differs from the prototype $p_k$ due to the different forward pass values %, i.e., $f(x) \neq f(p_{k})$, 
 %resulting in a different redistribution of relevances. %across feature map neurons. 
We refer the reader to \cite{montavon2019layer} for more information on LRP algorithms.

\subsection{Quantifying the Degree of Outlierness of a Sample}\label{outlier}

 Our proposed prototype-based architectures can naturally quantify the degree of outlierness of an input sample based on the sequence of sorted similarities %\cite{10.1007/978-3-642-13062-5_16, 10.1007/3-540-45681-3_2} 
 even though they are not trained primarily for outlier detection. We revisit the method in \cite{10.1007/978-3-642-13062-5_16} and generalize the confidence score in \cite{arik2020protoattend} to compute an outlier score based on the sum of similarity scores.
 We use the sum of the prototype similarity scores $u_{k}$ to compute the outlier score. While we see the main contribution in unifying two aspects of explanation and outlier detection, a smaller contribution lies in the different formulation of the prototype similarity scores $u_{k}$ with respect to the student architecture (refer to the output branch for outlier score in Figures \ref{head1}-\ref{head3}).

For each input sample $x$, we define the set of corresponding prototype similarity scores  as $U=\{u_{k} \ | \ 1 \leq k \leq K \}$. For all Head I and Head II architectures, the $u_{k}$ score is assigned as
$
u_{k}=z(p_k),
$
where $z(p_k)$ is the input to the linear layer $h(\cdot)$ as shown in Figures \ref{head1} and \ref{head2}. For 
Head III-A and Head III-B architectures, the $u_{k}$ score is computed as 
$u_{k}= \frac{1}{HW} \sum_{h=1,w=1}^{H,W} r_{h,w}(p_k).
$
For the Head III-C architecture, the $u_{k}$ score is computed as $ u_{k}=  \max_{h,w}(r_{h,w}(p_k)).$
%%uncomment this if include Head III-C and Head III-D architecture
%Finally, for the Head III-C and Head III-D architectures, the $u_{k}$ score is computed as $ u_{k}=  \max_{h,w}(r_{h,w}(x,p_k)).$ It is to be noted in this case that the substitution of the $r_{h,w}(x,p_k)$ with $r'_{h,w}(x,p_k)$ will produce the same $u_{k}$ score (refer to Section \ref{Network Architecture} for the formulation of $r'_{h,w}(x,p_k)$). 
With the placement of ReLU layers in the CNN, the elements of $U$ lie in the range of $[0,1]$, where a higher value indicates a larger similarity between the input sample $x$ and the prototype $p_k$. From the set $U$, top-$k'$ highest scores are selected regardless of the prototype class. 
For sample $x$, the outlier score $o(x)$ is defined as: 
\begin{equation}\label{outlier_eqn_score}
o= 1- \frac{1}{k'}\sum_{k=1}^{k'} u_k.
\end{equation}
Note that $u_k$ is in general no inner product, and thus, we cannot compute true metric distances and hence follow the idea in \cite{10.1007/978-3-642-13062-5_16}. In this formulation, we assume that the prototype similarity scores $u_k$ for outliers are smaller than the normal samples. The above formulation is the general case, which considers prototypes from all classes in the summation, unlike the formulation of the confidence score in \cite{arik2020protoattend}, which considers only the set of prototypes from the predicted class.
We show only the results using Eq. \eqref{outlier_eqn_score}, which gives better performance, and omit results using only prototypes from the same predicted class due to lack of space.
Larger outlier score $o$ indicates a larger anomaly.

\section{Experimental Settings}\label{exp_setup}
\textbf{Dataset.}
We used the LSUN \cite{yu2015lsun} dataset consisting of $10$ classes with an image size of $128\times128$ pixels, the PCam \cite{bejnordi2017diagnostic,veeling2018rotation} dataset consisting of $2$ classes with an image size of $96\times96$ pixels, and the Stanford Cars \cite{KrauseStarkDengFei-Fei_3DRR2013} dataset with image size $224\times224$ pixels, where we consider only samples in the first 50 cars classes instead of the total 196 classes due to time constraints.
For the LSUN dataset, we subsampled $10,000$ samples per class from the given training set to save training and computational time. We used the $80:20$ split for training and validation. The original validation split was used as the test set. For the Stanford Cars dataset, we used fewer number of training samples because we split 10\% of the given training set for each class to be our validation set, whereas the setup in \cite{chen2019looks} used all of the class samples in the given training set to train the model. We performed offline data augmentation following \cite{chen2019looks} on our training set and used the augmented training set for training.  Since each sample in the training set is augmented 29 times, using 10\% fewer base training samples will have a larger impact on the model performance, as the complete training set used for training consists of both augmented and base samples. In our setup, we used the performance on the validation set as an early training stopping criterion and to select the best model as opposed to using the training accuracy, cluster cost, and separation cost to stop training as done in \cite{chen2019looks}. 
Performance for the Stanford Cars dataset is reported on its official test set using only samples from the first 50 classes. For the PCam dataset, we used the given dataset splits as they are. We refrained from the common usage of CIFAR-10, MNIST, or Fashion MNIST classes, which are known to cluster very easily and offer little potential for aggregating spatial similarities due to low spatial resolutions. These datasets do not provide an accurate performance benchmark which we explain later in Section \ref{Prediction Performance}.

\textbf{LRP Perturbation.} We subsampled only $100$ samples per class from our LSUN test set and $500$ samples per class from our PCam test set to reduce computational time.
We set $\alpha=1.7$, $\beta=0.7$, and $\epsilon=1e-3$ for the LRP-$\alpha\beta$ and LRP-$\epsilon$ rules, respectively.

\textbf{Outlier Setup.} Three different types of outlier setups were used in our experiments. Setup A consists of the LSUN test set, which is labeled the normal sample set, and the test set from the Oxford Flowers 102 \cite{nilsback2008automated} dataset, which is labeled the outlier sample set. For Setup B, we created a synthetic outlier counterpart for each test sample in our test set (LSUN/PCam/Stanford Cars) by drawing random \textit{strokes} of thickness $M$ on the original image \cite{chong2020simple}. %More details regarding the creation of \textit{strokes} anomalies can be found in the work \cite{chong2020simple}. 
In Setup C, we created an outlier counterpart for each test sample by manipulating the color of the image, by increasing the minimum saturation and value components, and then randomly rotating the hue component. The manipulated image has abnormal colors.
Due to the large number of test samples in the PCam dataset, $1500$ samples per class were sampled from the test set for the outlier detection experiments. 

We showed a comparison with the common baseline method based on the maximum probability class \cite{hendrycks2016baseline}, the classic isolation forest (IF) \cite{liu2008isolation}, generalized ODIN (GODIN) \cite{hsu2020generalized}, and Outlier Exposure (OE) \cite{hendrycks2018deep}  to evaluate the potential of our prototype networks in distinguishing outliers from the normal sample distribution. %optimized for outlier detection. 
The outlier score for the baseline method \cite{hendrycks2016baseline} is the negative probability of the predicted class such that a sample with a smaller predicted class probability has a higher outlier score. For the training of GODIN layers, i.e.,  $g(x)$ and $h_i(x)$, we used a learning rate of $1e-3$, and we either fixed or fine-tuned the teacher network layers using a learning rate of $1e-4$ with an  early stopping training criterion using the performance on the in-distribution validation set. For the OE method, we fine-tuned the teacher network with a learning rate of $1e-3$  on one type of outlier and tested it on the other types of outliers. For fine-tuning with the out-of-distribution Oxford Flowers 102 dataset, we used the entire dataset comprising of training, validation, and test samples for training. For the hyperparameters of IF and other hyperparameters of GODIN and OE that are not stated here, we followed the hyperparameters used by the authors. We evaluated the outlier performance of the models using the area under the receiver operating characteristic (AUROC) curve, false positive rate at 95\% true positive rate (FPR95), and the area under the precision-recall (AUPR) curve metrics. We reported in Section S-II of the Supplementary Material both AUPRout and AUPRin metrics, where the former treats the outlier class as the positive class, while the latter treats the inlier class as the positive class.

\textbf{Model and Hyperparameter.} We employed ResNet-50 and ResNet-34 architectures \cite{he2016deep} pretrained on ImageNet as the CNN encoder. The former is used for the LSUN and PCam datasets, and the latter is used for the Stanford Cars dataset.
We used the SGD optimizer with momentum $0.9$ and weight decay of $1e-4$. For all head architecture networks, except Head III-B trained on the Stanford Cars dataset, the learning rates of $1e-3$ and $1e-4$ were used in the prototype network and the CNN encoder, respectively. A decay learning rate with a step size of $10$ epochs and $\gamma=0.1$ was also adopted.  To reduce overfitting for the Head III-B network on the Stanford Cars dataset, we used smaller learning rates of $5e-5$ and $1e-5$ for the prototype network and the CNN encoder, respectively, with learning rate decay every $20$ epochs and $\gamma=0.2$.  For the learning objectives, we set $\lambda_1=\lambda_2=1$ and $\lambda_3=0.1$. For the iterative prototype replacement algorithm, we set $p=30\%$ of the number of prototypes (Line 5 of Algorithm \ref{alg:protoselect}). We used PyTorch \cite{paszke2017automatic} for implementation. 

We compared our methods with the teacher network, ProtoDNN \cite{li2017deep}, ProtoPNet \cite{chen2019looks}, and the $\chi^2$-kernel SVM \cite{vedaldi2012efficient}. 
To control architectural effects, we used the same ResNet architecture pretrained on ImageNet as the CNN encoder for the aforementioned methods. %The teacher network used the same set of hyperparameters as our proposed network. 
We followed the hyperparameters used by the authors except for the learning rate of the encoder in ProtoDNN, which we set to a learning rate of $1e-5$.
%For the learning rate of the encoder in ProtoDNN, we used a learning rate of $1e-5$, while the remaining parts of the network used the learning rate of $1e-4$ as specified in the paper.
%We followed strictly the hyperparameters used by the authors for ProtoPNet and for the remaining hyperparameters of ProtoDNN. 
The $\chi^2$-kernel SVM used histogram CNN features pretrained on ImageNet. Since kernel SVMs do not scale to large datasets, we used a $\chi^2$ approximation kernel  \cite{vedaldi2012efficient} and the SGD classifier with hinge loss at an optimal learning rate and tolerance of $1e-4$. We performed a grid search on the validation set with the hyperparameters $\{1e-5, 1e-4,..., 1 \}$ to determine the regularization constant in the SGD classifier. We used $10$ prototypes per class in the LSUN and Stanford Cars experiments, and $20$ prototypes per class in the PCam experiments. %respectively, for all the prototype-based methods.

To ensure reproducibility, we provide the instructions to prepare the datasets and splits used in our experiments including the codes to create the synthetic outliers in GitHub\footnote{\url{https://github.com/pennychong94/Project_Towards-Scalable-and-Unified-Example-based-Explanation-and-OD}}.

\section{Experimental Results} \label{Results}
\subsection{Prediction Performance}\label{Prediction Performance}
We begin the comparison with the predictive aspect of our proposed architectures and the other methods on both datasets.
Based on Table \ref{performance-LSUN}, it can be observed that the proposed student architectures outperform the teacher network and are better than the other methods on the LSUN dataset. The Head III-B student network, which computes an attention map based on the maximum similarity between input and prototype, has the best classification performance among the other student networks. 
%comment this line if never use head III-C and head III-D architecture.
The Head III-C student network, which computes individual attention maps for the input and the prototype, has similar performance to Head III-B but is  computationally more expensive to train due to the additional attention map used. 
It can be observed that  ProtoDNN underperforms severely on the LSUN dataset.
%Since the network learns prototype vectors in the latent space, the vectors are decoded back to the input space for visualization. 
A manual inspection of the prototypes from ProtoDNN reveals that the prototypes do not  resemble any training samples and that the autoencoder used for prototype decoding fails to reconstruct the LSUN samples, resulting in  poor classification performance. This implies that the joint optimization for the layers in ProtoDNN performs well only on simple grayscale datasets such as MNIST and Fashion MNIST \cite{li2017deep} but is not generalizable to complex datasets such as LSUN. %The reconstruction performance of the autoencoder in ProtoDNN has an indirect impact on the classification performance of the network.
Our student network does not rely on autoencoders for prototype visualization, since the prototypes are actual training samples selected from the training set. 
%Our own observation with outlier detection on simpler datasets and the observation for ProtoDNN which generalize only on simpler datasets such as MNIST and Fashion MNIST, but not on complex datasets such as LSUN, has discouraged us from using the common but simple datasets in our evaluation to provide a more challenging performance benchmark.
On the basis of our own observation with outlier detection on simpler datasets and the inability of ProtoDNN to generalize to complex datasets such as LSUN, we refrain from using common and simple datasets such as MNIST and Fashion MNIST in our evaluation to provide a more challenging performance benchmark.
%The ProtoPNet which learns patch-wise prototype representation addressed the shortcoming of ProtoDNN by projecting the prototype vectors onto the nearest latent training patch from the same class and achieve reasonable classification performance on the LSUN dataset.

%%%%%%%%%%%%%%%%%%%%%%%%%%%%%%%
\begin{table}[ht!]
        \setlength\tabcolsep{25pt} % default value: 6pt
        \centering
       
        \caption{Classification performance on LSUN test set.}
        \label{performance-LSUN}
        \begin{tabular}{|c|c|}
        \hline
         \textbf{Model} & \textbf{ \ Acc. (\%)} \\ \hline
         \ $\chi^2$ SVM \cite{vedaldi2012efficient} \ & 76.8 \\
         \ ProtoDNN \cite{li2017deep} \ & 54.2 \\
         \ ProtoPNet \cite{chen2019looks} \ & 82.5 \\
        \ Teacher (\textit{baseline}) \ & 80.1 \\ \hhline{|=|=|}
         Student Head I \ & 82.9 \\ \cline{1-2} 
          Student Head II-A & 82.9 \\ %\cline{2-3} 
          Student Head II-B &  82.5 \\ \cline{1-2} 
         Student Head III-A \ & 82.8 \\ %\cline{2-3} 
         Student Head III-B \ &  \textbf{83.5}\\ %\cline{2-3} 
          Student Head III-C \ & 83.4 \\
          %Student Head III-D \ &  82.4\\
          \hline
        \end{tabular}
  
\end{table}
%%%pcam 
\begin{table}[ht!]
        \setlength\tabcolsep{12pt} % default value: 6pt
        \centering
     
        \caption{Classification performance on PCam test set.}
        
        \label{performance-Pcam}
        \begin{tabular}{|c|c|c|}
        \hline
        \textbf{Model} & \textbf{ \ Acc. (\%)} & \textbf{ \ AUROC \ } \\ \hline
        \ $\chi^2$ SVM \cite{vedaldi2012efficient} \ & 80.2  & 0.8875\\
        \ ProtoDNN \cite{li2017deep} \ & 76.7  & 0.8180\\
        \ ProtoPNet \cite{chen2019looks} \ &\textbf{83.4}  & 0.8822\\
        \ Teacher (\textit{baseline}) \ & 80.5  & 0.9142\\ \hhline{|=|=|=|}
         Student Head I \ & 83.3 & \textbf{0.9216} \\ \cline{1-3} 
          Student Head II-B & 82.5 & 0.9111 \\ \cline{1-3} 
         Student Head III-B \ & 81.0  & 0.9205 \\ \hline
        \end{tabular}

\end{table}
For the PCam and Stanford Cars datasets, as shown in Tables \ref{performance-Pcam} and \ref{performance-stanfordcar}, we show only one type of head per category. For the Head II and III categories, we show only performances of  Head II-B and Head III-B architectures as they are the general cases of Head II-A and Head III-A, respectively, and have lower computational complexity than the Head III-C architecture.   %%edit above if remove head III-C or add head III-D 
Based on Table \ref{performance-Pcam}, ProtoPNet and our student network with Head I architecture demonstrate equally good performance on the PCam dataset. The former achieves the highest test accuracy (only marginally better than Head I), and the latter achieves the best AUROC score. The PCam dataset consisting of only 2 classes and similar image statistics has smaller intraclass variations than the LSUN dataset with its 10 classes. In this case, Head I, which computes similarity using spatially pooled features, overfits less, thus performing better than more complex architectures such as Head II-B and Head III-B.

In Table \ref{performance-stanfordcar}, we compare the different head architectures only with the most competent state-of-the-art model, i.e., ProtoPNet. For the student networks trained on the Stanford Cars dataset, we observe similar performance trend as the student networks trained on PCam dataset since the Stanford Cars classification task is a fine-grained car model classification with significantly smaller intra- and interclass variations (similar to PCam samples) than the LSUN dataset. The architecture Head I has the best classification performance, surpassing the performance of ProtoPNet, while the attention-guided Head III-B architecture with higher complexity underperforms on this dataset. We used smaller learning rates for the Head III-B network on this dataset as the model overfits severely with higher learning rates as observed in our initial experiment. 
In an attempt to improve the  performance of the model on this dataset, we also investigated several Head III-B variants such as  a parameterized spatial attention and uniform attention variants but obtain unsatisfactory classification performance compared to the Head I and Head II-B networks. For more information, we refer the reader to Section S-V of the Supplementary Material. %We have also explored different variants of Head III-B such as a parameterized attention variant of Head III-B to be compared with the uniform attention variant that acts as a control experiment. 
Due to small sample variations in the fine-grained dataset and a larger number of learnable parameters in the Head III-B architecture (the additional parameters in the 1D-CNN layer), the model has a higher tendency to overfit compared to simpler architectures such as Head I and Head II-B. Since fine-grained classification is a task that involves distinguishing between very similar objects, the network may need to focus on a smaller region of the image sample. Thus, Head III-B can potentially be improved further by sampling small prototype patches from the training set, as inspired by the learning of prototype latent patches in ProtoPNet, which permits the attention map in Head III-B to be applied more precisely over a smaller target region. We note that there is a performance gap between the performance of ProtoPNet in Table \ref{performance-stanfordcar} and the performance reported in \cite{chen2019looks} (Table 1 of their Supplementary Material). The performance gap is due to our experimental setup,  which is different from \cite{chen2019looks} as described earlier in Section \ref{exp_setup}, and  the classification problem of the first $50$ classes that are naturally more challenging than the remaining Stanford Cars classes. For a complete discussion on this matter, we refer the reader to Section S-IV of the Supplementary Material.
%Since fine-grained classification task involves distinguishing between very similar objects, the network usually need to focus on a small region of the image sample and thus Head III-B may potentially improve by sampling small prototype patches as inspired by the learning of prototype latent patches in ProtoPNet.  which enables the attention map in Head III-B to be applied more precisely over a smaller target region.

% which are more prone to overfitting on simple datasets. %Similarly, ProtoDNN underperforms on the PCam dataset.
 For the remaining experiments using the PCam and Stanford Cars datasets, we evaluated only Head II-B and Head III-B for the Head II and Head III categories. 
 \begin{table}[ht!]
        \setlength\tabcolsep{25pt} % default value: 6pt
        \centering
       
        \caption{Classification performance on Stanford Cars test set using only samples from the first 50 classes. 
        The Head III-B network for this dataset is trained using lower learning rates than the other head architectures. Refer to Section \ref{exp_setup} for more details on the experimental setting. %i.e., learning rate $1e-4$ and $1e-5$ for the prototype network and the CNN encoder, respectively. 
        }
        \label{performance-stanfordcar}
        \begin{tabular}{|c|c|}
        \hline
         \textbf{Model} & \textbf{ \ Acc. (\%)} \\ \hline
         \ ProtoPNet \cite{chen2019looks} \ & 69.2 \\
        \ Teacher (\textit{baseline}) \ & 70.0 \\ \hhline{|=|=|}
         Student Head I \ & \textbf{72.1} \\ \cline{1-2} 
          Student Head II-B & 70.7  \\ \cline{1-2} 
         Student Head III-B \ & 60.6 \\ %\cline{2-3} 
          \hline
        \end{tabular}
  
\end{table}
 
%%%% include state-of-the
\begin{table}[ht!]
        \setlength\tabcolsep{25pt} % default value: 6pt
        \centering
       
        \caption{Performance of the state-of-the-art models compared to our teacher model on LSUN and PCam test sets. $^*$Our teacher network is trained on a subset of the full LSUN training set using only 8000 samples per class. The reported results for LSUN are the performances on the original validation split (consisting of 300 images per category) that we used as our test set. }
        \label{sota_perf}
        \begin{tabular}{|c|c|}
        \hline
         \textbf{LSUN Dataset} & \textbf{ \ Acc. (\%)} \\ \hline
        \ $^*$Teacher (\textit{baseline}) \ & 80.1 \\ 
         Context-CNN \cite{javed2017object}\ & 89.0 \\ 
         SIAT\_MMLAB \cite{wang2017knowledge} \ & \textbf{91.8}\\ \hline
         \textbf{PCam Dataset} & \textbf{ \ AUROC} \\\hline
          \ Teacher (\textit{baseline}) & 0.914 \\
          \ VF-CNN \cite{marcos2017rotation} & 0.951 \\
          \ G-CNN  \cite{bekkers2018roto,lafarge2021roto} & 0.968\\
          \ DSF-CNN \cite{graham2020dense} & \textbf{0.975}\\
          \hline
        \end{tabular}
  
\end{table}

We conclude that using prototype-based students generally results  in models that are equally competitive in prediction performance as the teacher network except for Head III-B on the Stanford Cars dataset. Employing the same underlying architecture and using soft labels from the teacher network to train the prototype networks allows unconstrained modeling of the original task. We note that our proposed models are built on top of a typical CNN model and thus have lower performance than more advance state-of-the-art methods that are curated to perform well on the task. To achieve performance comparable to Table \ref{sota_perf}, one may use any state-of-the-art model as the teacher network and use the soft labels and the same underlying architecture to train the prototype networks. We do not include the state-of-the-art performance on the Stanford Cars dataset in Table \ref{sota_perf} since we used only the first 50 car classes in our experiments due to time constraints.
%as they are the general cases of Head II-A and Head III-A, respectively, and have lower computational complexity than Head III-C architecture.
%%%%%%%%%%%%%%%%%%%%%%%%%%%%%%%%%%%%%%
\subsection{Top-$k$ Nearest Example and Pixelwise Explanation} 
We provide an explanation using the top-$k$ nearest prototypical examples from different classes and show the pixel evidence of similarity between the prototype and test sample.
We first compare the quality of the pixelwise LRP heatmap explanation generated by the different student architectures with the teacher network and then show examples of explanation in the two forms (refer to Section \ref{XAI}).
%for the student architecture with the best LRP explanation quality.
\begin{figure}[!htb]
\centering
  \includegraphics[height=1.7in, width=\linewidth]{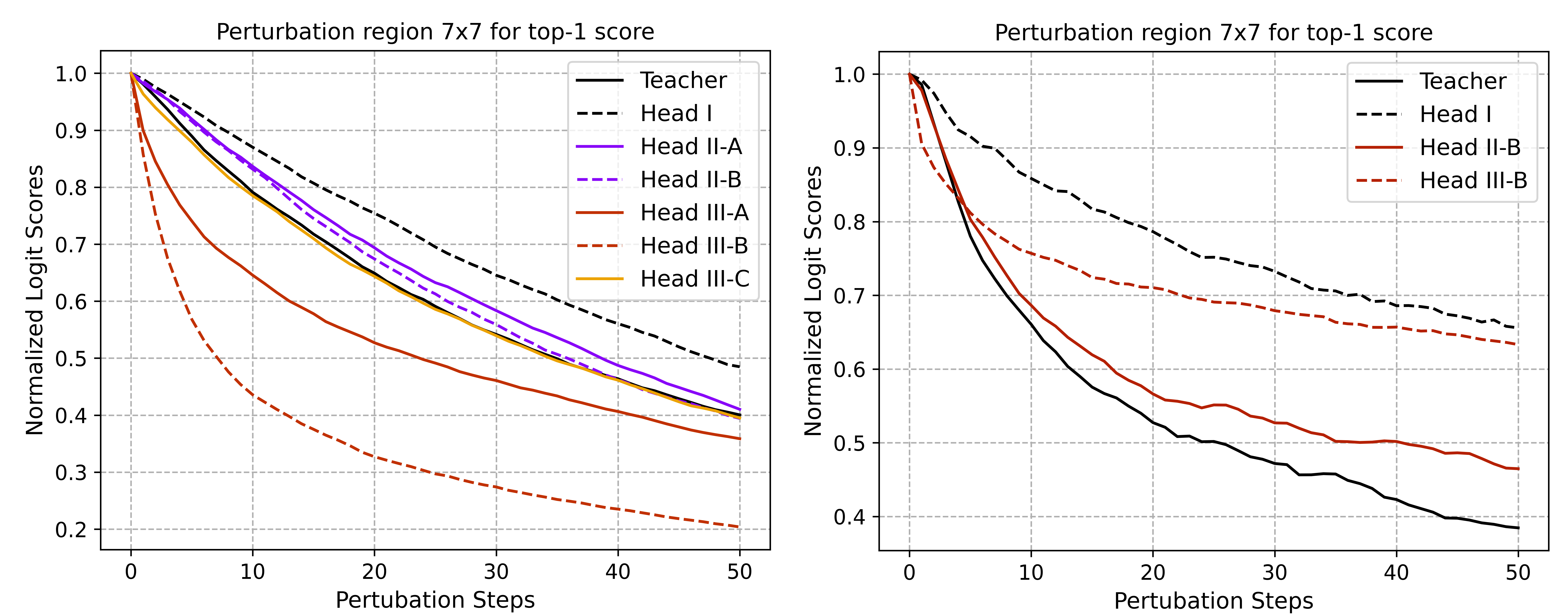}
\caption{LRP heatmap quality evaluation on LSUN (left) and PCam (right) datasets. The LRP heatmap perturbation for the student will use the LRP score computed between each test sample and its top-1 prototype.} \label{perturbfig} 
\end{figure}
\begin{figure*}[!htb]
\minipage{0.495\textwidth}
  \includegraphics[height=2.7in, width=\linewidth]{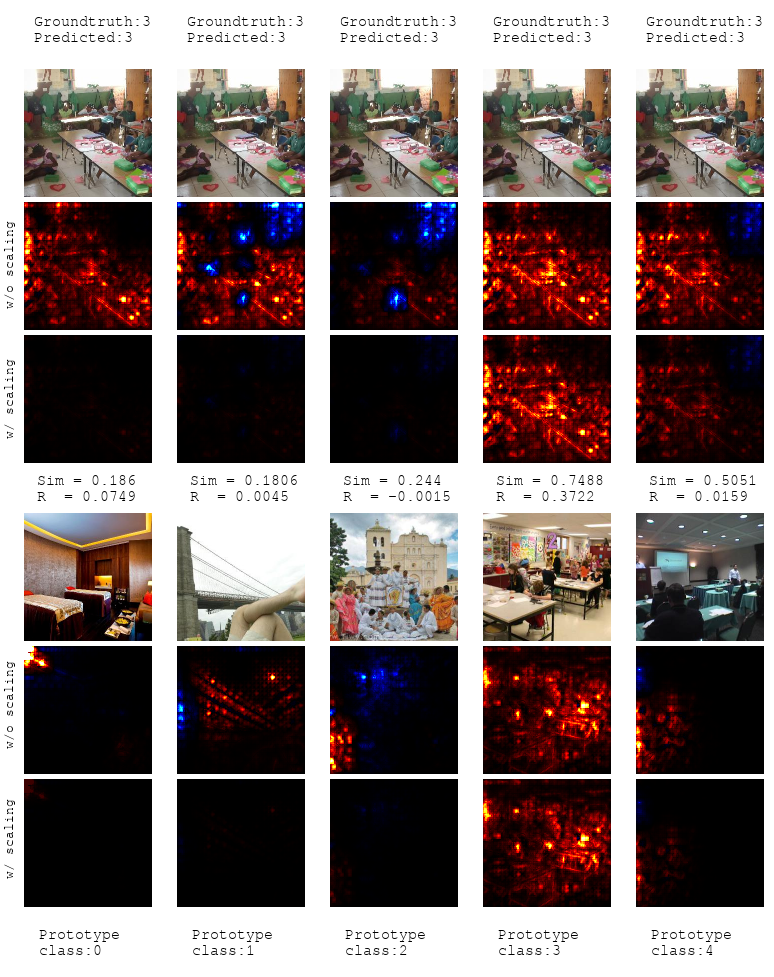}
\endminipage\hfill
\minipage{0.495\textwidth}
  \includegraphics[height=2.7in, width=\linewidth]{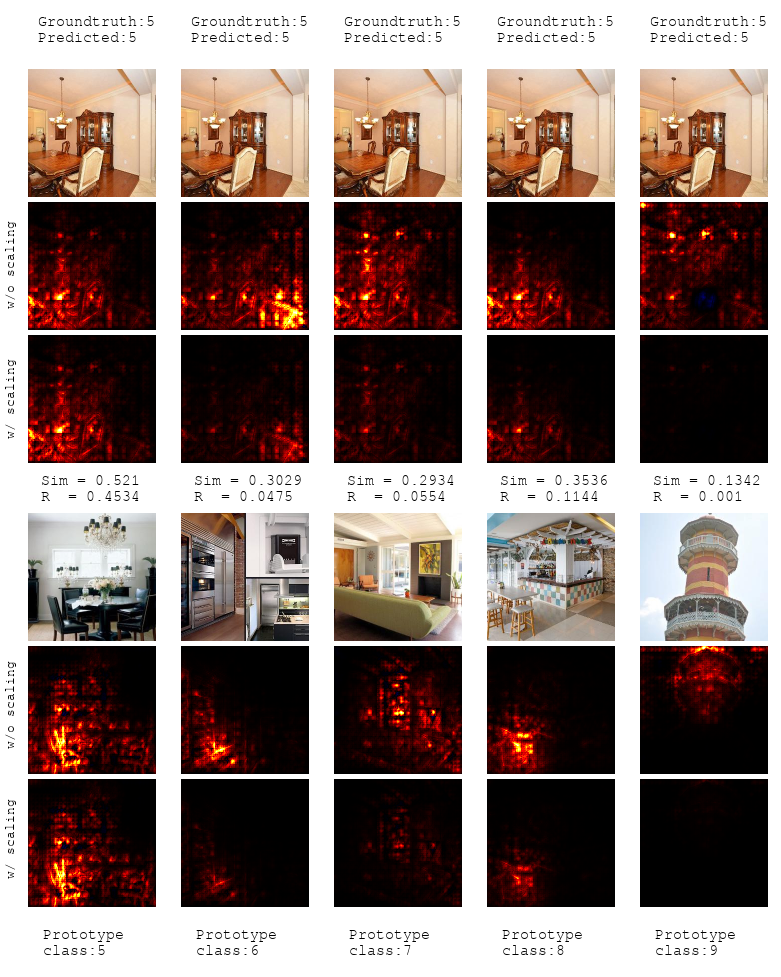}
\endminipage
\caption{LRP heatmaps  from Head III-B on the LSUN dataset for the prediction \textit{classroom} (left) and \textit{dining room} (right). Subplots in the upper part show heatmaps explaining the input sample, while those in the lower part show heatmaps explaining the shown prototype.
Each column corresponds to the heatmaps of an input-prototype pair with respect to the top-1 prototype of a specific class. The heatmaps in the second row of the upper and lower parts are unscaled, whereas the heatmaps in the third row are scaled such that the $(x,p_k)$ pair with lower $u_k$ score (written as sim score) will appear dimmer.} \label{lsun_cropped_heatmaps}
\end{figure*}
\begin{figure}[!htb]
\centering
\includegraphics[height=2.3in, width=\linewidth]{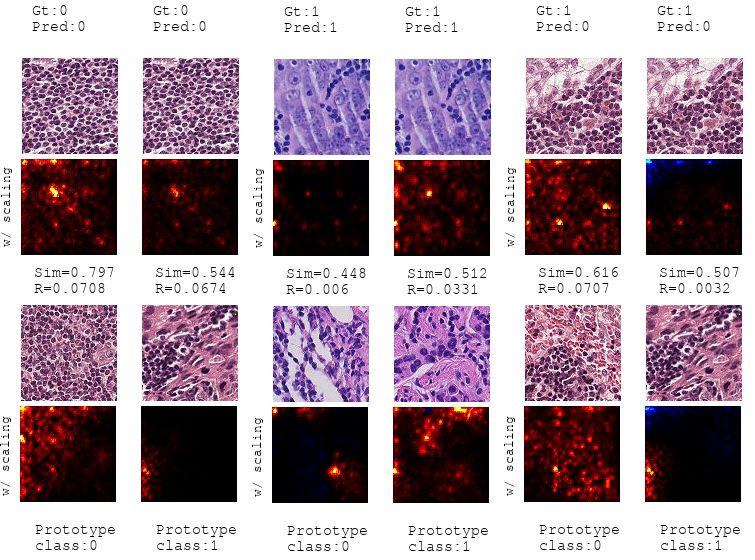}
\caption{LRP heatmaps from Head III-B on the PCam dataset.
The first two columns are for a true negative (tumor absent),
the third and fourth columns are for a true positive (tumor present), and the last two columns are for a false negative.
%The left, center, and right subplots are for a true negative (tumor absent), true positive (tumor present), and false negative predictions, respectively. **must update figure **
} \label{heatmaps_headIII-b_pcam}
\end{figure}
We assess the quality of the pixelwise LRP heatmap explanation for the network by performing a series of region perturbations in the input space of the test sample. The LRP heatmap comprises relevance scores associated with the degree of importance for the predicted class. Performing a perturbation on the image pixel associated with high relevance scores will destroy the network logit for the predicted class. In other words, we expect a sharp decrease in the logit score of the predicted class as the relevant features in the input sample are gradually removed in decreasing order of importance. We adopt the generalized region perturbation approach in \cite{samek2016evaluating}. %which is a greedy iterative procedure 
%that measures how the class encoded in the image disappears when information are progressively removed from the image $x$. 
We perform a series of perturbations starting from the most relevant region (based on the LRP relevance scores) to the least relevant region, as shown in Figure \ref{perturbfig}. A steep decrease suggests a good pixel explanation. 
%%% new ones

Based on the subplot for the LSUN dataset in Figure \ref{perturbfig}, the Head III-B student architecture with the best classification performance also has the best heatmap explanation, as shown by the largest decline in the logit scores. The attention modules in Head III-A and Head III-B show evidently better explanation qualities than the standard teacher network, as suggested by their respective curves. %The Head III-C which computes an additional attention map and Head II architectures present similar local explanation sensitivity as the teacher network, while Head I shows poorer explanation.
The Head I architecture shows a poorer explanation than the teacher network due to the early average pooling operation, which may have removed distinctive features that are useful in providing an informative pixel explanation. 
For the PCam dataset, the subplot suggests better heatmap quality for Head III-B architecture in the first 5 perturbation steps only and is overtaken by the teacher network and the Head II-B student network. 
Due to a lower diversity of objects in a patch (small cell types, stromal tissue and background) in the PCam dataset compared to the LSUN dataset, there may be a lesser need for attention weighting for explanation; thus, Head II-B shows better LRP explanation than its attention-augmented counterpart Head III-B when an increasing number of relevant regions are perturbed. Likewise, Head I architecture shows poor LRP explanation on the PCam dataset although it achieves the best classification performance in Table \ref{performance-Pcam}.  %The samples in the PCam dataset consists of a small number of cell types, stromal tissue and background. 
Generally, architectures with an attention mechanism provide better LRP explanation for datasets with large object diversity, while datasets with low object diversity need lesser attention weighting for explanation. In summary, Head II and Head III architectures produce heatmap quality comparable to the teacher network.

We show only LRP heatmaps from architecture with the best explanation (Head III-B) in Figures \ref{lsun_cropped_heatmaps} and \ref{heatmaps_headIII-b_pcam}. 
%For each input-prototype pair, we compute their respective LRP heatmaps. 
%For every test sample prediction, the student network presents explanation based on the nearest prototypical examples defined by $u_k$ and LRP heatmaps showing the  relevant regions in both test sample and the prototype, and these regions' impact to the membership of the predicted class in the form of positive and negative evidences.
We show pairs of heatmaps between the input sample and the top-1 nearest prototype (based on $u_k$ score)  for a few selected classes. 
The positive and negative evidence in the heatmaps are represented by red and blue pixels, respectively.
We obtain nonidentical input heatmaps (the columns in the top half of Figures \ref{lsun_cropped_heatmaps} and \ref{heatmaps_headIII-b_pcam}) when comparing the same input sample with different prototype samples due to the different similarity outputs computed and thus unequal relevances for each input-prototype pair.
%The positive and negative evidences supporting the network prediction in the LRP heatmap is represented by red and blue pixels, respectively.
%%%%%
%For each input sample, we compute one heatmap for the sample and for its top-1 nearest prototype from each class. 
In the left part of Figure \ref{lsun_cropped_heatmaps}, heatmaps for mostly all classes are dimmed out (the third row in the upper and lower part) after scaling due to their low similarity $u_{k}$ scores, except class $3$ (\textit{classroom}), which is the predicted class. The red pixels are the similar regions between the pair of input-prototype that has contributed positively to the prediction \textit{classroom}.
 The input heatmaps in the second row in columns 1, 4, and 5 look similar and they
correspond to the similarity of the input image (\textit{classroom}) to prototype examples belonging to the categories bedroom, classroom, and conference room, respectively. The categories bedroom, classroom, and conference room are all indoor scenes containing somewhat similar objects and indoor lighting. Since for each class we look at the nearest prototype to an image (rather than a randomly drawn prototype), we will likely find  similar sceneries for these three indoor categories in a large dataset such as LSUN. Thus, the input heatmaps that are computed based on looking at the nearest prototype for each category, appear to reflect a plausible similarity among these three similar indoor scene examples.
 
 In the lower right of Figure \ref{lsun_cropped_heatmaps} showing the top-1 prototypes from different classes,
%for the prediction \textit{dining room}, 
the regions that resemble furniture in the dining room, such as chairs, tables, and stools, generally lit up in the prototype heatmaps for the prediction \textit{dining room}.
They mainly correlate to the input regions with a dining table, chair, and chandelier, as shown in the upper right of Figure \ref{lsun_cropped_heatmaps}.

\begin{table*}[!ht]
\centering
\caption{Outlier detection performance reported using the \textbf{AUROC}$\vert$\textbf{FPR95} metric for LSUN networks. Higher ($\uparrow$) AUROC values indicate better outlier detection performance, whereas lower ($\downarrow$) FPR95 values indicate better outlier detection performance.}
\setlength\tabcolsep{3.5pt} % default value: 6pt
\label{outlier_lsun_alldata}
\begin{tabular}{|c|ccc|ccc|ccc|}
\hline
\multirow{1}{*}{\textbf{Model}} & \multicolumn{3}{c|}{\textbf{\begin{tabular}[c]{@{}c@{}}Setup A\\ Flowers\end{tabular}}} & \multicolumn{3}{c|}{\textbf{\begin{tabular}[c]{@{}c@{}}Setup B\\ LSUN strokes\_5\end{tabular}}} & \multicolumn{3}{c|}{\textbf{\begin{tabular}[c]{@{}c@{}}Setup C\\ LSUN altered color\end{tabular}}} \\ \hhline{|=|===|===|===|}
& \multicolumn{3}{c|}{\textit{\begin{tabular}[c]{@{}c@{}}Anomaly Score $^*$ \end{tabular}} } &  \multicolumn{3}{c|}{\textit{\begin{tabular}[c]{@{}c@{}}Anomaly Score $^*$\end{tabular}} } &  \multicolumn{3}{c|}{\textit{\begin{tabular}[c]{@{}c@{}}Anomaly Score $^*$ \end{tabular}} } \\   \cline{2-10}
IF \cite{liu2008isolation} & \multicolumn{3}{c|}{\begin{tabular}[c]{@{}c@{}} 0.678$\vert$0.661				\end{tabular}} &  \multicolumn{3}{c|}{\begin{tabular}[c]{@{}c@{}} 0.687$\vert$0.811	 \end{tabular}} &  \multicolumn{3}{c|}{\begin{tabular}[c]{@{}c@{}}0.482$\vert$0.936	 \end{tabular}} \\
$\textrm{Teacher}^1$ + GODIN \cite{hsu2020generalized} & \multicolumn{3}{c|}{\begin{tabular}[c]{@{}c@{}} 0.992$\vert$0.041 				\end{tabular}} &  \multicolumn{3}{c|}{\begin{tabular}[c]{@{}c@{}} 0.715$\vert$0.759	 \end{tabular}} &  \multicolumn{3}{c|}{\begin{tabular}[c]{@{}c@{}}\ 0.695$\vert$0.807	 \end{tabular}}\\
$\textrm{Teacher}^2$ + GODIN \cite{hsu2020generalized} & \multicolumn{3}{c|}{\begin{tabular}[c]{@{}c@{}} \textbf{0.995}$\vert$\textbf{0.019} 				\end{tabular}} &  \multicolumn{3}{c|}{\begin{tabular}[c]{@{}c@{}} 0.690$\vert$0.786	 \end{tabular}} &  \multicolumn{3}{c|}{\begin{tabular}[c]{@{}c@{}}0.692$\vert$0.824	 \end{tabular}}\\
\hhline{|=|===|===|===|}
& \multicolumn{3}{c|}{\textit{\begin{tabular}[c]{@{}c@{}}Max Prob. \end{tabular}} \cite{hendrycks2016baseline}} &  \multicolumn{3}{c|}{\textit{\begin{tabular}[c]{@{}c@{}}Max Prob. \end{tabular}} \cite{hendrycks2016baseline}} &  \multicolumn{3}{c|}{\textit{\begin{tabular}[c]{@{}c@{}}Max Prob. \end{tabular}} \cite{hendrycks2016baseline}} \\   \cline{2-10}
\ $\chi^2$ SVM \cite{vedaldi2012efficient} & \multicolumn{3}{c|}{\begin{tabular}[c]{@{}c@{}}0.921$\vert$0.313		\end{tabular}} &  \multicolumn{3}{c|}{\begin{tabular}[c]{@{}c@{}}0.742$\vert$0.737 \end{tabular}} &  \multicolumn{3}{c|}{\begin{tabular}[c]{@{}c@{}}0.691$\vert$0.820 \end{tabular}} \\
\ ProtoDNN \cite{li2017deep} & \multicolumn{3}{c|}{\begin{tabular}[c]{@{}c@{}}0.553$\vert$0.935			\end{tabular}} &  \multicolumn{3}{c|}{\begin{tabular}[c]{@{}c@{}}0.497$\vert$0.961 \end{tabular}} &  \multicolumn{3}{c|}{\begin{tabular}[c]{@{}c@{}}0.520$\vert$0.913 \end{tabular}} \\
\ ProtoPNet \cite{chen2019looks} & \multicolumn{3}{c|}{\begin{tabular}[c]{@{}c@{}}0.847$\vert$0.478		\end{tabular}} &  \multicolumn{3}{c|}{\begin{tabular}[c]{@{}c@{}}0.645$\vert$0.898 \end{tabular}} &  \multicolumn{3}{c|}{\begin{tabular}[c]{@{}c@{}}0.683$\vert$0.847 \end{tabular}} \\
\ Teacher (\textit{baseline})& \multicolumn{3}{c|}{\begin{tabular}[c]{@{}c@{}}0.934$\vert$0.284		\end{tabular}} &  \multicolumn{3}{c|}{\begin{tabular}[c]{@{}c@{}}0.677$\vert$0.800	 \end{tabular}} &  \multicolumn{3}{c|}{\begin{tabular}[c]{@{}c@{}}0.690$\vert$0.798	 \end{tabular}} \\
Teacher + OE \cite{hendrycks2018deep} (flowers)  & \multicolumn{3}{c|}{\begin{tabular}[c]{@{}c@{}} NA 				\end{tabular}} &  \multicolumn{3}{c|}{\begin{tabular}[c]{@{}c@{}} 0.679$\vert$0.812	 \end{tabular}} &  \multicolumn{3}{c|}{\begin{tabular}[c]{@{}c@{}}0.653$\vert$0.868	 \end{tabular}}\\
Teacher + OE \cite{hendrycks2018deep} ($\textrm{strokes}\_5$)  & \multicolumn{3}{c|}{\begin{tabular}[c]{@{}c@{}} 0.911$\vert$0.274 				\end{tabular}} &  \multicolumn{3}{c|}{\begin{tabular}[c]{@{}c@{}} NA	 \end{tabular}} &  \multicolumn{3}{c|}{\begin{tabular}[c]{@{}c@{}}0.710$\vert$0.821	 \end{tabular}}\\
Teacher + OE \cite{hendrycks2018deep} (color)  & \multicolumn{3}{c|}{\begin{tabular}[c]{@{}c@{}} 0.949$\vert$0.179 				\end{tabular}} &  \multicolumn{3}{c|}{\begin{tabular}[c]{@{}c@{}} 0.663$\vert$0.890	 \end{tabular}} &  \multicolumn{3}{c|}{\begin{tabular}[c]{@{}c@{}}NA	 \end{tabular}}\\
\hhline{|=|===|===|===|}
 & \textit{Top-1} & \textit{Top-20} & \textit{All proto.} & \textit{Top-1} & \textit{Top-20} & \textit{All proto.} & \textit{Top-1} & \textit{Top-20} & \textit{All proto.}\\ \cline{2-10}
\ Student Head I \ & 0.990$\vert$0.042 & 0.993$\vert$0.025  & 0.739$\vert$0.511  & 0.712$\vert$0.832 & 0.692$\vert$0.846 & 0.473$\vert$0.928  & 0.782$\vert$0.774 & \textbf{0.852}$\vert$\textbf{0.593}  & 0.630$\vert$0.734 \\ \hline
\ Student Head II-A \ & 0.970$\vert$0.129 & 0.965$\vert$0.151 & 0.786$\vert$0.632 & 0.764$\vert$0.719 & 0.738$\vert$0.799 &  0.611$\vert$0.858 & 0.809$\vert$0.661 & 0.835$\vert$0.631 & 0.710$\vert$0.826 \\
\ Student Head II-B \ & 0.972$\vert$0.151  & 0.984$\vert$0.082 & 0.986$\vert$0.068 & 0.763$\vert$0.749 & \textbf{0.765}$\vert$0.793 & 0.677$\vert$0.902 & 0.779$\vert$0.770 & 0.790$\vert$0.799 & 0.778$\vert$0.892 \\ \hline
\ Student Head III-A \ & 0.922$\vert$0.315 & 0.943$\vert$0.221 & 0.885$\vert$0.328 & 0.726$\vert$0.748 & 0.735$\vert$\textbf{0.701}  & 0.683$\vert$0.772 & 0.672$\vert$0.822 & 0.691$\vert$0.766  & 0.636$\vert$0.853 \\
\ Student Head III-B \ & 0.855$\vert$0.657 & 0.906$\vert$0.542 & 0.909$\vert$0.434 & 0.692$\vert$0.837 & 0.702$\vert$0.842 & 0.718$\vert$0.766 & 0.674$\vert$0.852 & 0.707$\vert$0.834 & 0.731$\vert$0.772\\
\ Student Head III-C \ & 0.970$\vert$0.118 & 0.972$\vert$0.071 & 0.914$\vert$0.434 & 0.726$\vert$0.769 & 0.727$\vert$0.772 & 0.509$\vert$0.890 & 0.763$\vert$0.763 & 0.786$\vert$0.690 & 0.680$\vert$0.804 \\
%\ Student Head III-D \ & 0.980 & 0.988  & 0.859  & 0.706 & 0.682 & 0.390  & 0.740  & 0.758  &  0.599 \\ 
\hline
\multicolumn{10}{l}{\begin{tabular}[c]{@{}c@{}}	\footnotesize{$^*$ The anomaly detection algorithm returns an anomaly score for every input sample, and such an algorithm is not used for classification.}\\
\end{tabular}}\\
\multicolumn{10}{l}{\begin{tabular}[c]{@{}c@{}}	\footnotesize{$^1$ The parameters in the trained layers of the teacher network are fixed, and optimization involves only the GODIN layers, i.e., $g(x)$ and $h_i(x)$.}\\
\end{tabular}}\\
\multicolumn{10}{l}{\begin{tabular}[c]{@{}c@{}}	\footnotesize{$^2$ The parameters in the trained layers of the teacher network are fine-tuned with a learning rate that is an order smaller than the learning}\\
\end{tabular}}\\
\multicolumn{10}{l}{\begin{tabular}[c]{@{}c@{}}	\footnotesize{rate for the GODIN layers, i.e., $g(x)$ and $h_i(x)$.}\\
\end{tabular}}\\
%is not meant for classification.
\multicolumn{10}{c}{\begin{tabular}[c]{@{}c@{}}		\end{tabular}}
\end{tabular}
\end{table*}

%%%%%%%%%%%%%%%%%%%%%%%%%%%%%%%%%%%%%%%%%%%%
\begin{table*}[!ht]
\centering
\caption{Outlier detection performance reported using the \textbf{AUROC}$\vert$\textbf{FPR95} metric for PCam networks. Higher ($\uparrow$) AUROC values indicate better outlier detection performance, whereas lower ($\downarrow$) FPR95 values indicate better outlier detection performance.}

\label{outlier_pcam_alldata}
\begin{tabular}{|c|ccc|ccc|}
\hline
\multirow{1}{*}{\textbf{Model}} & \multicolumn{3}{c|}{\textbf{\begin{tabular}[c]{@{}c@{}}Setup B\\ PCam strokes\_5\end{tabular}}} & \multicolumn{3}{c|}{\textbf{\begin{tabular}[c]{@{}c@{}}Setup C\\ PCam altered color\end{tabular}}} \\ \hhline{|=|===|===|}
& \multicolumn{3}{c|}{\textit{\begin{tabular}[c]{@{}c@{}}Anomaly Score $^*$ \end{tabular}} } &  \multicolumn{3}{c|}{\textit{\begin{tabular}[c]{@{}c@{}}Anomaly Score $^*$\end{tabular}} }  \\   \cline{2-7} 
IF \cite{liu2008isolation} & \multicolumn{3}{c|}{\begin{tabular}[c]{@{}c@{}} \textbf{0.723}$\vert$\textbf{0.712}	\end{tabular}} &  \multicolumn{3}{c|}{\begin{tabular}[c]{@{}c@{}}0.580$\vert$0.754 \end{tabular}} \\
$\textrm{Teacher}^1$ + GODIN \cite{hsu2020generalized} & \multicolumn{3}{c|}{\begin{tabular}[c]{@{}c@{}} 0.696$\vert$0.824	\end{tabular}} &  \multicolumn{3}{c|}{\begin{tabular}[c]{@{}c@{}}0.585$\vert$0.967 \end{tabular}} \\
$\textrm{Teacher}^2$ + GODIN \cite{hsu2020generalized} & \multicolumn{3}{c|}{\begin{tabular}[c]{@{}c@{}} 0.579$\vert$0.929	\end{tabular}} &  \multicolumn{3}{c|}{\begin{tabular}[c]{@{}c@{}}0.788$\vert$0.857 \end{tabular}} \\
\hhline{|=|===|===|} 
& \multicolumn{3}{c|}{\textit{\begin{tabular}[c]{@{}c@{}}Max Prob. \end{tabular}} \cite{hendrycks2016baseline}} &  \multicolumn{3}{c|}{\textit{\begin{tabular}[c]{@{}c@{}}Max Prob. \end{tabular}} \cite{hendrycks2016baseline}} \\   \cline{2-7}
\ $\chi^2$ SVM \cite{vedaldi2012efficient} & \multicolumn{3}{c|}{\begin{tabular}[c]{@{}c@{}}0.521$\vert$0.938		\end{tabular}} &  \multicolumn{3}{c|}{\begin{tabular}[c]{@{}c@{}}0.519$\vert$0.933 \end{tabular}} \\  
%%ProtoDNN \cite{li2017deep}
\ ProtoDNN \cite{li2017deep} & \multicolumn{3}{c|}{\begin{tabular}[c]{@{}c@{}}0.441$\vert$0.976			\end{tabular}} &  \multicolumn{3}{c|}{\begin{tabular}[c]{@{}c@{}}0.087$\vert$0.991 \end{tabular}} \\ 
\ ProtoPNet \cite{chen2019looks} & \multicolumn{3}{c|}{\begin{tabular}[c]{@{}c@{}}0.514$\vert$0.930			\end{tabular}} &  \multicolumn{3}{c|}{\begin{tabular}[c]{@{}c@{}}0.450$\vert$0.654 \end{tabular}} \\ 
\ Teacher (\textit{baseline}) & \multicolumn{3}{c|}{\begin{tabular}[c]{@{}c@{}}0.527$\vert$0.905		\end{tabular}} &  \multicolumn{3}{c|}{\begin{tabular}[c]{@{}c@{}}0.326$\vert$0.910			\end{tabular}} \\ 
\ Teacher + OE \cite{hendrycks2018deep} ($\textrm{strokes}\_5$) & \multicolumn{3}{c|}{\begin{tabular}[c]{@{}c@{}}NA		\end{tabular}} &  \multicolumn{3}{c|}{\begin{tabular}[c]{@{}c@{}}0.703$\vert$0.640			\end{tabular}} \\ 
\ Teacher + OE \cite{hendrycks2018deep} (color) & \multicolumn{3}{c|}{\begin{tabular}[c]{@{}c@{}}0.561$\vert$1.00		\end{tabular}} &  \multicolumn{3}{c|}{\begin{tabular}[c]{@{}c@{}}NA			\end{tabular}} \\ 
\hhline{|=|===|===|}
 & \textit{Top-1} & \textit{Top-20} & \textit{All proto.} & \textit{Top-1} & \textit{Top-20} & \textit{All proto.} \\ \cline{2-7}
\ Student Head I \ & 0.576$\vert$0.817 & 0.574$\vert$0.816 & 0.511$\vert$0.960 & 0.350$\vert$0.822 & 0.358$\vert$0.853 &  0.820$\vert$0.427\\ \
\ Student Head II-B \ & 0.678$\vert$0.753 & 0.661$\vert$0.730 & 0.636$\vert$0.882 & 0.377$\vert$0.929 & 0.394$\vert$0.870 & \textbf{0.823}$\vert$\textbf{0.325}\\ \
\ Student Head III-B \ & 0.687$\vert$0.778 & 0.631$\vert$0.829 & 0.541$\vert$0.835 & 0.329$\vert$0.908  & 0.249$\vert$0.955 &  0.224$\vert$0.959 \\
%\multicolumn{7}{l}{\begin{tabular}[c]{@{}c@{}}	%\footnotesize{$^*$ The anomaly detection algorithm returns an anomaly score for every input sample and such}
%\end{tabular}}\\
%\multicolumn{7}{l}{\begin{tabular}[c]{@{}c@{}}	\footnotesize{ \ \ algorithm is not used for classification. }
%\end{tabular}}\\
%
\hline
\multicolumn{7}{l}{\begin{tabular}[c]{@{}c@{}}	\footnotesize{$^*$ The anomaly detection algorithm returns an anomaly score for every input sample, and such an algorithm is not used for}\\
\end{tabular}}\\
\multicolumn{7}{l}{\begin{tabular}[c]{@{}c@{}}	\footnotesize{classification.}\\
\end{tabular}}\\
\multicolumn{7}{l}{\begin{tabular}[c]{@{}c@{}}	\footnotesize{$^1$ The parameters in the trained layers of the teacher network are fixed, and optimization involves only the GODIN layers,}\\
\end{tabular}}\\
\multicolumn{7}{l}{\begin{tabular}[c]{@{}c@{}}	\footnotesize{i.e., $g(x)$ and $h_i(x)$.}\\
\end{tabular}}\\
\multicolumn{7}{l}{\begin{tabular}[c]{@{}c@{}}	\footnotesize{$^2$ The parameters in the trained layers of the teacher network are fine-tuned with a learning rate that is an order smaller}\\
\end{tabular}}\\
\multicolumn{7}{l}{\begin{tabular}[c]{@{}c@{}}	\footnotesize{than the learning rate for the GODIN layers, i.e., $g(x)$ and $h_i(x)$.}\\
\end{tabular}}\\
%is not meant for classification.
\multicolumn{7}{c}{\begin{tabular}[c]{@{}c@{}}		\end{tabular}}
\end{tabular}

\end{table*}
%%%%%%%%%%%%%%%%%%%%%%% For stanford cars
\begin{table*}[!ht]
\centering
\caption{Outlier detection performance reported using the \textbf{AUROC}$\vert$\textbf{FPR95} metric for Stanford Cars networks. Higher ($\uparrow$) AUROC values indicate better outlier detection performance, whereas lower ($\downarrow$) FPR95 values indicate better outlier detection performance.}

\label{outlier_stanfordcar_alldata}
\begin{tabular}{|c|ccc|ccc|}
\hline
\multirow{1}{*}{\textbf{Model}} & \multicolumn{3}{c|}{\textbf{\begin{tabular}[c]{@{}c@{}}Setup B\\ Stanford Cars strokes\_5\end{tabular}}} & \multicolumn{3}{c|}{\textbf{\begin{tabular}[c]{@{}c@{}}Setup C\\ Stanford Cars altered color\end{tabular}}} \\ \hhline{|=|===|===|}
& \multicolumn{3}{c|}{\textit{\begin{tabular}[c]{@{}c@{}}Anomaly Score $^*$ \end{tabular}} } &  \multicolumn{3}{c|}{\textit{\begin{tabular}[c]{@{}c@{}}Anomaly Score $^*$\end{tabular}} }  \\   \cline{2-7} 
IF \cite{liu2008isolation} & \multicolumn{3}{c|}{\begin{tabular}[c]{@{}c@{}} \textbf{0.874}$\vert$\textbf{0.533}	\end{tabular}} &  \multicolumn{3}{c|}{\begin{tabular}[c]{@{}c@{}} 0.854$\vert$0.531 \end{tabular}} \\
$\textrm{Teacher}^1$ + GODIN \cite{hsu2020generalized} & \multicolumn{3}{c|}{\begin{tabular}[c]{@{}c@{}} 0.622$\vert$0.940	\end{tabular}} &  \multicolumn{3}{c|}{\begin{tabular}[c]{@{}c@{}}0.503$\vert$0.994 \end{tabular}} \\
$\textrm{Teacher}^2$ + GODIN \cite{hsu2020generalized} & \multicolumn{3}{c|}{\begin{tabular}[c]{@{}c@{}} 0.466$\vert$0.987	\end{tabular}} &  \multicolumn{3}{c|}{\begin{tabular}[c]{@{}c@{}}0.529$\vert$0.955 \end{tabular}} \\
\hhline{|=|===|===|} 
& \multicolumn{3}{c|}{\textit{\begin{tabular}[c]{@{}c@{}}Max Prob. \end{tabular}} \cite{hendrycks2016baseline}} &  \multicolumn{3}{c|}{\textit{\begin{tabular}[c]{@{}c@{}}Max Prob. \end{tabular}} \cite{hendrycks2016baseline}} \\   \cline{2-7}
\ ProtoPNet \cite{chen2019looks} & \multicolumn{3}{c|}{\begin{tabular}[c]{@{}c@{}}0.728$\vert$1.000	\end{tabular}} &  \multicolumn{3}{c|}{\begin{tabular}[c]{@{}c@{}}0.792$\vert$0.592 \end{tabular}} \\ 
\ Teacher (\textit{baseline}) & \multicolumn{3}{c|}{\begin{tabular}[c]{@{}c@{}}0.760$\vert$0.725 \end{tabular}} &  \multicolumn{3}{c|}{\begin{tabular}[c]{@{}c@{}}0.789$\vert$0.635	\end{tabular}} \\ 
\ Teacher + OE \cite{hendrycks2018deep} ($\textrm{strokes}\_5$) & \multicolumn{3}{c|}{\begin{tabular}[c]{@{}c@{}}NA		\end{tabular}} &  \multicolumn{3}{c|}{\begin{tabular}[c]{@{}c@{}}\textbf{0.998}$\vert$\textbf{0.100}		\end{tabular}} \\ 
\ Teacher + OE \cite{hendrycks2018deep} (color) & \multicolumn{3}{c|}{\begin{tabular}[c]{@{}c@{}}0.851$\vert$0.569		\end{tabular}} &  \multicolumn{3}{c|}{\begin{tabular}[c]{@{}c@{}}NA			\end{tabular}} \\ 
\hhline{|=|===|===|}
 & \textit{Top-1} & \textit{Top-20} & \textit{All proto.} & \textit{Top-1} & \textit{Top-20} & \textit{All proto.} \\ \cline{2-7}
\ Student Head I \ & 0.815$\vert$0.706 & 0.808$\vert$0.723 & 0.627$\vert$0.844 & 0.838$\vert$0.675 & 0.859$\vert$0.551 &  0.725$\vert$0.729\\ \
\ Student Head II-B \ & 0.818$\vert$0.687 & 0.799$\vert$0.770 & 0.393$\vert$0.953 & 0.810$\vert$0.662 & 0.782$\vert$0.741 &  0.265$\vert$0.975 \\
\ Student Head III-B \ & 0.700$\vert$0.857 & 0.661$\vert$0.899& 0.551$\vert$0.890 & 0.913$\vert$0.361& 0.931$\vert$0.307 &  0.903$\vert$0.346 \\
\hline
\multicolumn{7}{l}{\begin{tabular}[c]{@{}c@{}}	\footnotesize{$^*$ The anomaly detection algorithm returns an anomaly score for every input sample, and such an algorithm is not used for}\\
\end{tabular}}\\
\multicolumn{7}{l}{\begin{tabular}[c]{@{}c@{}}	\footnotesize{classification.}\\
\end{tabular}}\\
\multicolumn{7}{l}{\begin{tabular}[c]{@{}c@{}}	\footnotesize{$^1$ The parameters in the trained layers of the teacher network are fixed, and optimization involves only the GODIN layers,}\\
\end{tabular}}\\
\multicolumn{7}{l}{\begin{tabular}[c]{@{}c@{}}	\footnotesize{i.e., $g(x)$ and $h_i(x)$.}\\
\end{tabular}}\\
\multicolumn{7}{l}{\begin{tabular}[c]{@{}c@{}}	\footnotesize{$^2$ The parameters in the trained layers of the teacher network are fine-tuned with a learning rate that is an order smaller}\\
\end{tabular}}\\
\multicolumn{7}{l}{\begin{tabular}[c]{@{}c@{}}	\footnotesize{than the learning rate for the GODIN layers, i.e., $g(x)$ and $h_i(x)$.}\\
\end{tabular}}\\
%is not meant for classification.
\multicolumn{7}{c}{\begin{tabular}[c]{@{}c@{}}		\end{tabular}}
\end{tabular}

\end{table*}
%%%%%%%%%%%%%%%%%%%%%%%%%%%%%%%%%%
%For the explanation on the PCam dataset in Figure \ref{heatmaps_headIII-b_pcam}, we showin the first two columns the explanation for a true negative, the third and fourth columns the explanation for a true positive, and finally the explanation for a false negative in the last two columns.
For the explanation on the PCam dataset in Figure \ref{heatmaps_headIII-b_pcam}, we show
explanations for a true negative, a true positive, and a false negative.
%The LRP heatmaps on the top row are for the  input sample, while the bottom heatmaps are for the top-1 nearest prototype of the shown class.
The true negative prediction shows positive relevance scores (in red) on dense clusters of lymphocytes (dark regular nuclei), which is evidence supporting  the absence of tumors. %The closest prototype from the same negative class (class 0) is also an image with dense regular nuclei that resembles the input sample.
For the false negative prediction, the input, which is a positive sample (as indicated by the ground truth), resembles more of the negative class prototype (class 0) rather than the positive class prototype (class 1) in the red pixel areas.
One usually looks at  the nearest prototype to the test sample for explanation in addition to assessing the heatmap computed for the test sample. In the PCam  case,  the interpretation of the prototype with its LRP heatmap  requires some domain knowledge from the person, such as a pathologist who is trained to recognize certain morphological structures.
\begin{figure}[!htb]
\centering
  \includegraphics[height=1.5in, width=2in]{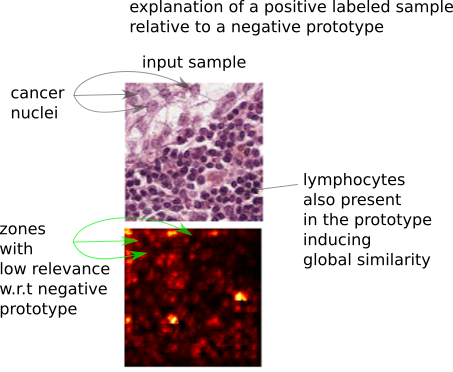}
  \caption{Input sample and its heatmap for a false negative prediction extracted from the first two rows and the fifth column of Figure \ref{heatmaps_headIII-b_pcam}. The input heatmap is computed based on its similarity with the predicted negative class prototype, as shown in the fifth column of Figure \ref{heatmaps_headIII-b_pcam}.}
  \label{extracted_input_heatmap_FN}
\end{figure}
In Figure \ref{extracted_input_heatmap_FN}, which depicts the false negative explanation extracted from Figure \ref{heatmaps_headIII-b_pcam}   for the input sample  with respect to a negative class prototype, one can see in  the upper left of the first row (displaying the input sample)  a few faint likely cancer nuclei. The heatmap shown in Figure \ref{extracted_input_heatmap_FN} shows very low heatmaps score in these cancer nuclei locations (almost black) and high heatmap scores in their direct surrounding embedding tissue matrix.
 Moreover, one sees high heatmap scores in some of the black dots, which are tissue invading lymphocytes (TiLs) also present in the prototype, and for which some but not all of them seem to be similar to the TiLs in the prototype (Figure \ref{heatmaps_headIII-b_pcam}, column 5, row 3).
 By looking at the prototype and its heatmap in the lower part of Figure \ref{heatmaps_headIII-b_pcam}, column 5, one can see that the similarities are found in a part of the tissue invading lymphocytes and a part of the embedding stromal tissue matrix.

The explanations in Figures \ref{lsun_cropped_heatmaps} and \ref{heatmaps_headIII-b_pcam} allow domain experts to easily validate the reasoning process of the neural networks by verifying that the nearest prototypical training samples are  correct and the contributing features are relevant to the predicted class. By analyzing contradictory cases of prediction and  explanation, domain experts understand the model limitation and are able to make an informed judgment. We have included more LRP explanation examples in Section S-III of the Supplementary Material.

The prototype similarity scores $u_k$, that are used to identify the top-$k$ nearest prototypical examples can also be employed to gauge the explanation confidence of a given input sample. The optimal value of $k$ that we choose to use for explanation depends on the task at hand. We can consider an explanation based on prototypical examples to be of sufficient confidence if the top-$k$ nearest prototypes (i.e., top-$k$ highest $u_k$ score) from the predicted class are responsible for a high fixed ratio (e.g., 90\%) of the prediction mass; thus, these prototypes can be considered to have sufficiently high similarity scores. The top-$k$ nearest prototypes are considered to have insufficient similarity scores if their total score is below this fixed ratio.

Alternatively, one can also interpret the similarity scores in another manner.
We consider the set of top-$k$ nearest prototypes from the predicted class to be too small if the removal of the prototype with the smallest similarity score would change the prediction label such that it differed from the prediction with the full set of prototypes. This is related to the principle of pertinent positives \cite{dhurandhar2018explanations}, but performed on the prototype level rather than the level of subsets of an input sample.

\begin{figure}[!htb]
\centering
  \includegraphics[height=1.7in, width=\linewidth]{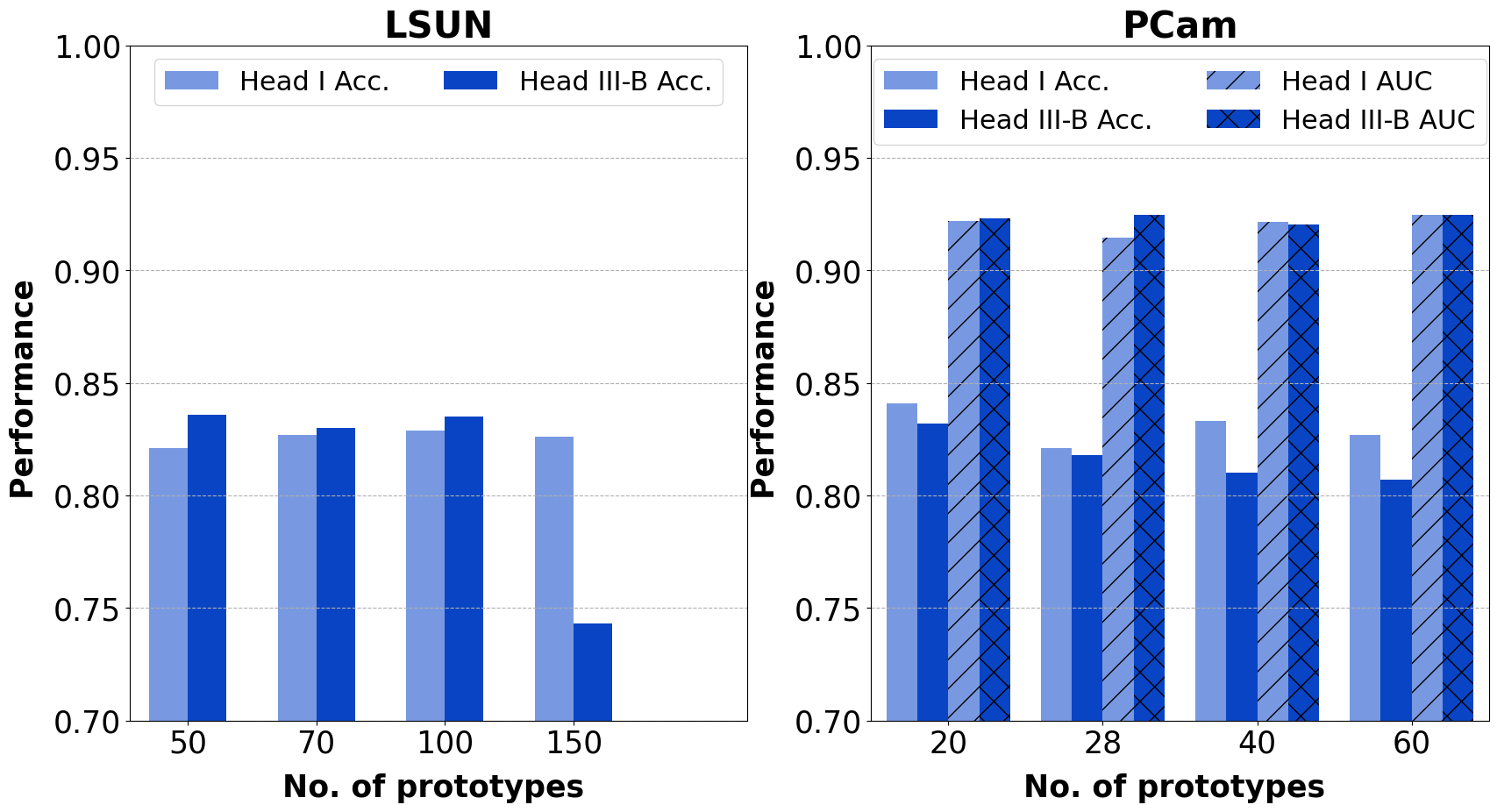}
\caption{Impact on performance using different numbers of prototypes. The x-axis is the total  number of prototypes from all classes.} \label{varynumproto} 
\end{figure}
\begin{table}[ht!]
        \centering
        \caption{Classification performance without and with pruning.}
        \label{pruning-LSUN}
        \begin{tabular}{|c|c|c||c|c|}
        \hline
         \multirow{2}{*}{\textbf{Model}} & \multirow{2}{*}{\textbf{Prune}}& \textbf{LSUN}& \multicolumn{2}{c|}{\begin{tabular}[c]{@{}c@{}}\textbf{PCam}\end{tabular}}\\ \cline{3-5}
         & & \textbf{ Acc. (\%)} & \textbf{ Acc. (\%)} & \textbf{AUROC} \\ \hline
          \multirow{2}{*}{Head I} & \xmark & \textbf{82.9} & \textbf{83.3} & \textbf{0.9216}\\ 
          &\cmark & 82.7 & 82.6 & 0.9177\\ \hline      
          \multirow{2}{*}{Head II-B} &\xmark & 82.5 &  82.5& 0.9111 \\ 
          &\cmark & \textbf{82.6} & \textbf{83.3} & 0.9111\\ \hline   
          \multirow{2}{*}{Head III-B} &\xmark& \textbf{83.5} & 81.0 & 0.9205\\ 
          &\cmark & 83.2 & \textbf{81.8} & \textbf{0.9244}\\ \hline   
        \end{tabular}
\end{table}
%For the false negative prediction, the explanation extracted from the network shows that the input sample which is a positive sample (as indicated by the ground truth) resembles more of the negative class prototype (class 0) rather than the positive class prototype (class 1), particularly in the areas that lit up in red.
%The true negative prediction highlights positive scores on dense clusters of lymphocytes (the dark regular nuclei). Not shown due to lack of space, we confirmed that the closest prototype is an image of tissue-invading lymphocytes. 
%For the false negative, the heatmap focuses on structures resembling cancer nuclei, which however are outside of the $32\times32$ window which defines the label for a $96\times96$ patch of the PCam dataset. Thus, the heatmap is plausible given that the label considers only the $32\times32$ center.
%%%%
\begin{table*}[!ht]
\centering
\setlength\tabcolsep{2.5pt} % default value: 6pt
\caption{Outlier detection performance reported using the AUROC metric for networks without and with pruning. %For the pruned networks, $30\%$ of the total prototypes and their relevant neurons are pruned.
}
\label{outlier_lsun_alldata_pruning}
\begin{tabular}{|c|c|ccc|ccc|ccc||ccc|ccc|}
\hline
\multirow{2}{*}{\textbf{Model}} &\multirow{2}{*}{\textbf{Prune}}& \multicolumn{9}{c||}{\textbf{\begin{tabular}[c]{@{}c@{}}LSUN\end{tabular}}} & \multicolumn{6}{c|}{\textbf{\begin{tabular}[c]{@{}c@{}}PCam\end{tabular}}} \\ \cline{3-17}
& & \multicolumn{3}{c|}{\textbf{\begin{tabular}[c]{@{}c@{}}Setup A\\ Flowers\end{tabular}}} & \multicolumn{3}{c|}{\textbf{\begin{tabular}[c]{@{}c@{}}Setup B\\ LSUN strokes\_5\end{tabular}}} & \multicolumn{3}{c||}{\textbf{\begin{tabular}[c]{@{}c@{}}Setup C\\ LSUN altered color\end{tabular}}} & \multicolumn{3}{c|}{\textbf{\begin{tabular}[c]{@{}c@{}}Setup B\\ PCam strokes\_5\end{tabular}}} & \multicolumn{3}{c|}{\textbf{\begin{tabular}[c]{@{}c@{}}Setup C\\ PCam altered color\end{tabular}}}\\ 
\hhline{|=|=|===|===|===||===|===|}
 & & \textit{Top-1} & \textit{Top-20} & \textit{All proto.} & \textit{Top-1} & \textit{Top-20} & \textit{All proto.} & \textit{Top-1} & \textit{Top-20} & \textit{All proto.}&  \textit{Top-1} & \textit{Top-20} & \textit{All proto.} &  \textit{Top-1} & \textit{Top-20} & \textit{All proto.}\\ \cline{3-17}

\multirow{2}{*}{Head I} \ & \xmark &  0.990 & \textbf{0.993}  & 0.739  & \textbf{0.712} & 0.692 & 0.473  & 0.782 & \textbf{0.852}  & 0.630 & 0.576 & 0.574 & 0.511 & 0.350 & 0.358 &  \textbf{0.820}  \\
 & \cmark & 0.986  & 0.982  & 0.704  & 0.701 & 0.626  & 0.477  & 0.773 &  0.821 & 0.635 & \textbf{0.579} & 0.506 & 0.467 & 0.367 & 0.323 & 0.584\\
\hline
\multirow{2}{*}{Head II-B} \ & \xmark &   0.972  & 0.984 & \textbf{0.986} & 0.763 & 0.765 & 0.677 & 0.779 & 0.790 & 0.778 &  0.678 & 0.661 & 0.636 & 0.377 & 0.394 & 0.823\\
 & \cmark & 0.968 &  0.983 & 0.977  & 0.749 & 0.765  & 0.672  & 0.757 &  \textbf{0.812} & 0.777 &  0.673 & \textbf{0.701}  & 0.676 & 0.443 & 0.552 & \textbf{0.825} \\
\hline
\multirow{2}{*}{Head III-B} \ & \xmark &   0.855 & 0.906 & 0.909 & 0.692 & 0.702 & 0.718 & 0.674 & 0.707 & 0.731 & \textbf{0.687} & 0.631 & 0.541 & \textbf{0.329}  & 0.249 &  0.224 \\
 & \cmark & 0.821 & \textbf{0.948}  & 0.933  & 0.689 & \textbf{0.754}  & 0.742  & 0.677 & \textbf{0.780}   &  0.774 & 0.641 & 0.622  & 0.623 &  0.234 & 0.211 & 0.257 \\\hline
\end{tabular}
\end{table*}
\subsection{Detecting Outliers for Additional Validation}
%Besides providing an example-based explanation, 
The prototype student networks can naturally quantify the degree of outlierness using the set of similarity scores.
The GODIN method shows the best outlier detection performance in the flowers setup, as shown in Table \ref{outlier_lsun_alldata}, due to significant differences in the data distribution between the LSUN and flowers datasets. GODIN does not perform well on more challenging outlier setups such as strokes and altered color setups in which anomalies are created via the manipulation of normal samples. %resulting in the creation of anomalous samples with similar data distribution as the normal samples.

%%%%
The Head I student architecture in Table \ref{outlier_lsun_alldata}  achieves the second-best performance for the flowers setup and top performance for the altered color setup using the top-$20$ prototype similarity scores $u_k$. Since the image statistics of the flowers dataset and the altered color anomalies are substantially different from the normal LSUN samples, the use of early spatial pooling before computing similarity in Head I does not remove distinctive features of the anomalies but is effective in detecting outliers while offering the least potential for overfitting.  For the strokes outlier setup, Head II-B %which computes similarity before the spatial pooling operation 
generally has the best detection performance, as shown in Table \ref{outlier_lsun_alldata} and Table I of the Supplementary Material. The placement of the spatial pooling operation before the similarity operation in Head I may have removed traces of strokes on the image, while Head II-B retains the traces of strokes during the similarity computation. 
%which leads to the successful detection of strokes anomalies. %Although Head III-B student network gives the best classification accuracy and LRP explanation on the LSUN dataset, 
The Head III-B architecture has a higher architectural complexity than Head I and Head II and is thus less robust to outliers and prone to overfitting. %Generally, we found that the flowers setup is easier than the strokes and altered color setups as suggested by the average detection scores. 
We show the outlier performance of LSUN networks under the AUPRout and AUPRin metrics in Section S-II of the Supplementary Material due to space constraints.

We report detection only on the more challenging setups (strokes and altered color as suggested by the average detection scores in Table \ref{outlier_lsun_alldata}) for the PCam and Stanford Cars datasets. 

In Table \ref{outlier_pcam_alldata}, the IF method generally achieves the best detection results on the strokes setup. Likewise, the student architectures with similarity layers before averaging (Head II-B and Head III-B) perform better in detecting strokes artifacts on the PCam dataset, which is consistent with the LSUN networks. For the altered color setup, as shown in Table \ref{outlier_pcam_alldata}, student networks with lower architectural complexity, such as Head I and Head II-B architectures, perform well on the altered color setup compared to Head III-B with higher architectural complexity.
The Head II-B and Head I networks using all prototypes are equally competent on the PCam altered color setting.
The outliers created using the PCam dataset in this setting are easier to be detected than those created using the LSUN dataset due to the significant differences between the color distribution of outliers and normal PCam samples. 
%Although the Head II-B network does not perform as well as Head I network in the LSUN altered color setup, the Head II-B network performs marginally better than Head I on the PCam altered color setting. The outliers created using the PatchCam dataset in this setting are easier to be detected as compared to the LSUN dataset due to the significant differences between the color distribution of outliers and normal PCam samples. 
%In this easier setting, the similarity operation in Head II-B that is performed on unpooled features does not overfit and retains distinctive outlier features. 
We also observe that the ProtoDNN underperforms on the altered color setup, as a majority of the outliers have a higher maximum class probability than the normal samples. %which contradicts the assumption in \cite{hendrycks2016baseline}.
We show the outlier performance of PCam networks under the AUPRout and AUPRin metrics in Section S-II of the Supplementary Material due to space constraints.

For the Stanford Cars experiments in Table \ref{outlier_stanfordcar_alldata}, we also observe that the IF model performs best in the strokes setup, similar to the PCam experiments. The competing OE method achieves comparable performance to IF on the Stanford Cars strokes setup and superior performance on the altered color setup. We deduce that exposing the network to an outlier set  during training when the in-distribution set has low variability among samples can be highly beneficial to the model, as it is able to learn a more conservative concept of inliers easily as opposed to a setting with highly varied in-distribution samples. %\changemarker{We also observe that the Head III-B network with poor classification performance in Table \ref{performance-stanfordcar} does not perform as well as Head II-B in the strokes outlier setup as opposed to its performance in Table \ref{outlier_pcam_alldata}.} 
We also observe that the Head III-B network with poor classification performance in Table \ref{performance-stanfordcar} performs better than the other head architectures in the altered color setup but not in the strokes setup as opposed to its performance on PCam outliers in Table \ref{outlier_pcam_alldata}.
Due to lack of space, we report the outlier performance of Stanford Cars networks under the AUPRout and AUPRin metrics in Section S-II of the Supplementary Material.
%In the Stanford Cars strokes setup, Head I and Head II-B have similar performance using top-1 or top-20 prototypes. Since the strokes outlier can appear on any part of the image, the Head I and Head II-B models that learned fine-grained features of car models may focus more on certain regions of the sample that may not contain traces of strokes.

In conclusion, except for the PCam altered color setup, the use of 20 prototypes for Head II and Head III architectures generally shows reasonable outlier performances, and are mostly better than or comparable to the IF, GODIN, and OE methods. %although they are not optimized for outlier detection. 
This supports our argument for unified prototype-based models with generic head architectures that are interpretable and robust to outliers. 
%This supports our  suggestion of unified prototype-based models combined with the prototype selection and the proposed heads. 

\section{Discussion}\label{Discussion}
\subsection{Varying Number of Prototypes}
We investigate the impact on classification performance using different numbers of prototypes. We use an equal number of prototypes per class. We discuss only the performance for student networks with simple (Head I) and complex (Head III-B) architectures due to a lack of space.
Based on Figure \ref{varynumproto}, only a marginal change in performance can be seen for the simple Head I architecture on the LSUN dataset, whereas a huge decline in performance  can be seen for the more complex Head III-B architecture when using a large number of prototypes (150 prototypes) due to its larger tendency to overfit. %At a small number of prototypes ($\leq$ 100), the change in classification performance is minimal for Head III-B on the LSUN dataset. 
Likewise, only slight performance differences are reported for the PCam dataset with different numbers of prototypes. For the relatively simple PCam dataset, using a small number of prototypes (20 prototypes) is sufficient and less likely to overfit, as shown by the high test accuracy for both architectures. %A minimal change in performance is observed for the AUROC metric with larger number of prototypes (60 prototypes) reporting slightly higher AUROC on the PCam dataset. 
Hence, using a small number of prototypes for simple datasets and complex network architectures is a good practice.

\subsection{Pruning of Prototypes and Relevant Neurons}\label{Pruning of Prototypes and Relevant Neurons}
Since the prototype importance weight $\mathcal{M}$ defines the importance %or the contribution 
of the $K$ prototypes in the network prediction, we can prune the set of prototypes with the least weight in $\mathcal{M}$ to evaluate the impact on classification and outlier performances. We do not evaluate the LRP explanation of the pruned networks, as the general architecture of the networks remains largely unchanged. %except for the lesser number of neurons in the linear layer and the convolutional-1D layer (for Head III).
We prune $p=30\%$ (the same as $p$ used in the iterative  prototype replacement algorithm) of the prototypes and the relevant neurons, and then we fine-tune the network.   Alternatively, one can set the pruning factor based on the classification performance on the validation set.
%Alternatively, one can also optimize the number of prototypes to prune by zero-ing out the relevant prototype vectors and measuring the classification performance of the network on the validation set before fine-tuning.
%%%%merge table (outlier detection pruning)

Based on Table \ref{pruning-LSUN}, we observe marginal changes in the classification performance after pruning. The pruned network with Head II-B shows a slight increase in accuracy for both datasets. %, whereas Head III-B shows a small gain in performance only on the PCam dataset. 
The outlier detection performance of the pruned networks is reported in Table \ref{outlier_lsun_alldata_pruning}. The pruned networks exhibit a larger change in the outlier performance compared to the prediction accuracy. For the LSUN dataset, the student network with the Head III-B architecture consistently performs better with pruning, as indicated by the significant gain in the detection performance. Since the pruning of prototypes and relevant neurons reduces the size of the network, the Head III-B network overfits less %on the training data 
after fine-tuning and is more robust to outliers. We do not observe the same benefit of pruning for Head III-B on the PCam outlier setups. For the strokes setup, an evident gain in performance can be observed for the Head II-B network trained on the PCam dataset. A smaller gain is seen for the Head II-B network in the altered color setup for both datasets. Thus, one may consider pruning uninformative prototypes if the network is prone to overfitting. % and their relevant neurons %to reduce the number of parameters in the network 
%if the network is prone to overfitting. 
%%%%%%%%%%%%%%%%%%%%%%

\section{Conclusion}
Various student architectures that compute prototype-based predictions were investigated for their suitability to provide example-based explanations and detect outliers.
The selection of prototypical examples from the training set via the iterative prototype replacement method guaranteed meaningful explanation. The prototype similarity scores can also be used naturally to quantify the outlierness of a sample. %However, there is a trade-off between the quality of the generated explanations and the outlier detection performance. 
However, the architecture with the best LRP explanation did not achieve the best outlier detection performance. %For future work, we will explore the application of the proposed network for other tasks. 

% or


% \appendices

% % you can choose not to have a title for an appendix
% % if you want by leaving the argument blank
% \section{}

\section*{Acknowledgment}
This research is supported by both ST Engineering Electronics and NRF, Singapore, under its Corporate Lab @ University Scheme (Title: STEE Infosec-SUTD Corporate Laboratory) and the Research Council of Norway, via the SFI Visual Intelligence (Grant Number: 309439).

\ifCLASSOPTIONcaptionsoff
  \newpage
\fi



%\bibliographystyle{IEEEtran}


% \end{thebibliography}
\bibliographystyle{IEEEtran}
\bibliography{icprworkshop}
\vskip -2\baselineskip plus -1fil
% \begin{IEEEbiographynophoto}{Penny Chong}
% is a Ph.D. student at the Singapore University of Technology and Design (SUTD), under the supervision of Alexander Binder and Ngai-Man Cheung. She received a B.Sc. (Hons) in Applied Mathematics with Computing from Universiti Tunku Abdul Rahman (UTAR), Malaysia in 2016. Her research interests include machine learning, explainable AI and its applications.
\begin{IEEEbiographynophoto}{Penny Chong}
is a research scientist at IBM Research, Singapore.
She received her Ph.D. in Information Systems Technology and Design from Singapore University of Technology and Design (SUTD) in 2021, under the supervision of Alexander Binder and Ngai-Man Cheung. Her research interests include machine learning and explainable AI and its applications.
\end{IEEEbiographynophoto}
\vskip -2\baselineskip plus -1fil
\begin{IEEEbiographynophoto}{Ngai-Man Cheung}
is an associate professor at the Singapore University of Technology and Design (SUTD). He received his Ph.D. degree  in Electrical Engineering from the University of Southern California (USC), Los Angeles, CA, in 2008. His research interests include image and signal processing, computer vision and AI.
\end{IEEEbiographynophoto}
\vskip -2\baselineskip plus -1fil
\begin{IEEEbiographynophoto}{Yuval Elovici}
is the director of the Telekom Innovation Laboratories at Ben-Gurion University of the Negev (BGU), head of the BGU Cyber Security Research Center and a professor in the Department of Software and Information Systems Engineering at BGU. %He holds B.Sc. and M.Sc. degrees in Computer and Electrical Engineering from BGU and a Ph.D. in Information Systems from Tel-Aviv University.
He holds a Ph.D. in Information Systems from Tel-Aviv University.
His primary research interests are computer and network security, cyber security, web intelligence, information warfare, social network analysis, and machine learning. %He is the co-founder of the start-up Morphisec. 
%His primary research interests are computer and network security, cyber security, web intelligence, information warfare, social network analysis, and machine learning. He also consults professionally in the area of cyber security and is the co-founder of Morphisec, a start-up company that develop innovative cyber security mechanisms that relate to moving target defense.
\end{IEEEbiographynophoto}
\vskip -2\baselineskip plus -1fil
\begin{IEEEbiographynophoto}{Alexander Binder}
is an associate professor at the University of Oslo (UiO). He received a Dr.rer.nat. from Technische Universit\"at Berlin, Germany in 2013. He was an assistant professor at the Singapore University of Technology and Design (SUTD) from 2015 to 2020. %He has participated in Pascal VOC and ImageCLEF competitions. 
His research interests include explainable deep learning among other topics.
\end{IEEEbiographynophoto}

% \end{IEEEbiography}


% % insert where needed to balance the two columns on the last page with
% %\newpage


%\vfill


\end{document}